%% file: main_arxiv.tex
\documentclass{article} % For LaTeX2e
\usepackage{arxiv,times}

\input{math_commands.tex}

\renewcommand{\eqref}[1]{(\ref{#1})}
\usepackage{graphicx}

\usepackage{amsmath}

\usepackage{hyperref}
\usepackage{url}
\newtheorem{theorem}{Theorem}
\newtheorem{lemma}{Lemma}
\newtheorem{assumption}{Assumption}
\newtheorem{definition}{Definition}

\usepackage[T1]{fontenc}
\usepackage{xcolor}
\usepackage{tabularray}
\iclrfinalcopy

\title{Stochastic heavy ball with \\ Polyak step size and Armijo line search: \\a general convergence analysis}

\author{Jiawei Zhang, Qitan Shi \& Yuantao Gu\thanks{Corresponding author: \texttt{gyt@tsinghua.edu.cn}.} \\
Tsinghua University \\
}

\begin{document}

\maketitle

\input{Sections/0.abs}

\input{Sections/1.intro}

\input{Sections/2.background}

\input{Sections/3.general_convergence}

\input{Sections/4.exact_convergence_beyond}

\input{Sections/5.numerical_results}

\input{Sections/6.conclusion}

\bibliography{iclr2027_conference}
\bibliographystyle{iclr2027_conference}

\clearpage

\appendix
\input{Appendix/a.proof_of_lemmas}

\input{Appendix/b.proof_of_theorems}

\input{Appendix/c.proof_of_theorems_diminishing}

\input{Appendix/d.more_related_works}

\input{Appendix/e.numerical_proof_of_concepts}

\end{document}

%% file: math_commands.tex
\usepackage{amsmath,amsfonts,bm}

\def\eqref#1{equation~\ref{#1}}
\def\1{\bm{1}}

\DeclareMathAlphabet{\mathsfit}{\encodingdefault}{\sfdefault}{m}{sl}
\SetMathAlphabet{\mathsfit}{bold}{\encodingdefault}{\sfdefault}{bx}{n}

%% file: Sections/0.abs.tex
\begin{abstract}
Polyak step size (PS) and Armijo line search (ALS) have received increasing attention in stochastic optimization, with encouraging empirical performance and theoretical guarantees.
However, their convergence theory for stochastic heavy ball (SHB) methods remains limited.
In this work, we develop a unified convergence analysis for SHB equipped with PS and ALS.
To this end, we introduce a modified Armijo rule that closely parallels the Polyak step size, together with a decoupling analysis that isolates the historical dependence induced by momentum.
For SHB with standard PS and ALS, we establish expected convergence for strongly convex, convex, and non-convex objectives without interpolation or restrictive conditions on the momentum parameter.
Under interpolation or strong growth, we further strengthen the results to almost sure rates and last-iterate convergence.
Moreover, for general settings beyond interpolation, we prove almost sure convergence to the exact optimum or to stationarity for SHB with diminishing variants of PS and ALS.
These results provide a more comprehensive theoretical view of Polyak step size and Armijo line search for stochastic heavy ball methods.
\end{abstract}

%% file: Sections/1.intro.tex
\section{Introduction}
\label{sec:intro}

We consider the stochastic optimization problem
\begin{align}
\label{eq:stochastic_optimization_problem}
    \min_{\boldsymbol{\theta}} F(\boldsymbol{\theta})
    = \mathbb{E}_{x \sim p_x} f(\boldsymbol{\theta};{x}).
\end{align}
where $\boldsymbol{\theta}$ denotes the model parameter to be optimized, and $x$ denotes a random data sample drawn from the data distribution $p_x$.
This formulation captures a broad class of fundamental problems in machine learning and related fields~\citep{shalev2009stochastic, bottou2018optimization, krizhevsky2012imagenet, he2016deep, ren2015faster, lin2017focal, vaswani2017attention}.
A widely used method for solving such problems is the stochastic heavy ball (SHB) method~\citep{polyak1964some, sutskever2013importance, gadat2018stochastic}, whose iteration is given by
\begin{align}
\label{eq:shb}
    \boldsymbol{\theta}_{k+1}
    =
    \boldsymbol{\theta}_k
    -
    \gamma_k \nabla f(\boldsymbol{\theta}_k;x_k)
    +
    \beta(\boldsymbol{\theta}_k-\boldsymbol{\theta}_{k-1}),
\end{align}
where $x_k$ is sampled i.i.d. from $p_x$, $\beta\in[0,1)$ is the momentum parameter, and $\gamma_k$ is the step size at iteration $k$.
We initialize $\boldsymbol{\theta}_{-1}=\boldsymbol{\theta}_0$.
While the momentum term often improves practical convergence~\citep{sutskever2013importance, gitman2019understanding}, the resulting historical dependence makes SHB substantially harder to analyze than standard stochastic gradient descent (SGD)~\citep{yang2016unified, loizou2020momentum, sebbouh2021almost, ganesh2023does}.

In this paper, we focus on SHB with adaptive step sizes, specifically Polyak step size (PS)~\citep{polyak1987introduction} and Armijo line search (ALS)~\citep{armijo1966minimization}.
For SGD without momentum, PS and ALS have shown strong empirical performance and enjoy solid theoretical guarantees~\citep{sls, berrada2020training, SPS}.
%
% Motivated by the success of PS and ALS in SGD, several recent works have explored adaptive step-size rules for momentum-based methods~\citep{ALR-SMAG, MomSPS, SGDM-APS, fan2023msl, lapucci2025convergence, lapucci2026effectively}.
Motivated by the success in SGD, several recent works have explored variants of PS and ALS for momentum-based optimization algorithms~\citep{ALR-SMAG, fan2023msl, MomSPS, SGDM-APS, lapucci2025convergence, lapucci2026effectively}.
Nevertheless, a general convergence theory for SHB with PS and ALS remains incomplete. Existing guarantees typically apply only to restricted settings, such as convex objectives or functions satisfying the PL condition, and often rely on interpolation or additional assumptions on the momentum parameter $\beta$~\citep{MomSPS, lapucci2025convergence, lapucci2026effectively}.

We address this gap by developing a more general convergence analysis for SHB with PS and ALS.
To analyze the two step-size rules in a unified way, we first introduce a modified Armijo rule whose key step-size properties align with those of PS under the SHB iteration.
We then provide two technical lemmas that decouple the momentum-dependent terms in SHB: one controls the difference between successive iterates, and the other controls the change in the objective value along the SHB trajectory.
Building on these ingredients, we derive convergence rates for SHB with PS and ALS under strongly convex, convex, and general non-convex objectives.
%
% These results do not impose the interpolation assumption or restrictions on the momentum parameter $\beta$, and thus better reflect the large-momentum regimes commonly used in practice.
These results do not impose the interpolation assumption and apply for any fixed $\beta\in[0, 1)$, without an additional explicit upper bound on the momentum hyperparameter.
Under interpolation or strong growth, we further strengthen these results to almost sure rates and last-iterate convergence in both convex and non-convex settings, which is particularly relevant to overparameterized deep learning models.
Moreover, beyond interpolation, we analyze SHB with diminishing variants of PS and ALS and establish almost sure exact-convergence guarantees covering both convex and general non-convex objectives.

We summarize our main results in Table~\ref{tab:convergence_summary},
and our contributions are as follows:
\begin{itemize}
    \item We develop a unified convergence framework for SHB with PS and ALS, using a modified Armijo rule to align their key step-size properties and two decoupling lemmas to isolate the dependence between current iterates and historical gradients induced by momentum.
    
    \item We establish expected convergence guarantees for strongly convex, convex, and non-convex objectives without interpolation, for any fixed momentum parameter $\beta\in[0,1)$ under suitable step-size conditions. Under interpolation or strong growth, we further obtain almost sure last-iterate convergence in objective value or gradient norm, respectively.
    
    % \item Beyond interpolation, we establish almost sure convergence rates for SHB equipped with diminishing PS and ALS and a summable momentum schedule. These results guarantee convergence of the weighted-average objective value to the exact optimum for convex objectives and of the best-iterate gradient norm to zero for general non-convex objectives.
    \item Beyond the interpolation or strong growth, we establish almost-sure convergence for SHB equipped with diminishing variants of PS and ALS under a summable momentum schedule. Specifically, we obtain almost-sure rates for the weighted-average objective gap in the convex setting and the best-iterate squared gradient norm in the general non-convex setting.
\end{itemize}

\input{tables/table_results_summary}

%% file: tables/table_results_summary.tex
\begin{table}[!t]
\centering
\caption{Summary of the main results in this paper.
$\bar{\boldsymbol{\theta}}_k$ and
$\bar{\boldsymbol{\theta}}_k^{\rm w}$ denote the
uniform and weighted averages of the iterates, respectively.
$S_k$ denotes the
cumulative upper bound of the effective step sizes.
Under interpolation or strong growth, we further establish last-iterate convergence.
For the diminishing variants, the corresponding convergence rates can be sharpened from big-\(\mathcal{O}\) to little-\(o\) bounds.\\}
\label{tab:convergence_summary}
\begingroup
\hypersetup{hidelinks}
\small
\begin{tblr}{
  width=\linewidth,
  colspec={X[1.15,l] X[1.15,l] X[2.7,l] Q[c]},
  cells={valign=m},
  colsep=5pt,
  rowsep=1.3pt,
  row{1}={font=\bfseries},
  row{2,6,9}={bg=gray!10,font=\bfseries},
  cell{2,6,9}{1}={c=4}{l},
  hline{1,Z}={0.8pt},
  hline{2}={0.4pt},
}
Objective & Noise condition & Guarantee & Theorem \\

Standard PS / ALS: convergence in expectation
& & & \\

Strongly convex
& $\sigma^2<\infty$
& $\mathbb E\|{\boldsymbol{\theta}}_k-{\boldsymbol{\theta}}^\star\|^2
   =\mathcal O(r^k+\sigma^2)$
& \ref{thm:strongly_convex_general} \\

Convex
& $\sigma^2<\infty$
& $\mathbb E[F(\bar{\boldsymbol{\theta}}_k)-F({\boldsymbol{\theta}}^\star)]
   =\mathcal O(k^{-1}+\sigma^2)$
& \ref{thm:convex_general} \\

Non-convex
& Weak growth
& $\min_{0\le m<k}\mathbb E\|\nabla F({\boldsymbol{\theta}}_m)\|^2
   =\mathcal O(k^{-1}+\delta)$
& \ref{thm:nonconvex_general} \\

Standard PS / ALS: almost sure convergence
& & & \\

Convex
& {Interpolation}
& {$F(\bar{\boldsymbol{\theta}}_k)-F({\boldsymbol{\theta}}^\star)=\mathcal O(k^{-1}),\qquad \text{a.s.}$}
& \ref{thm:convex_as} \\

Non-convex
& {Strong growth}
& {$\min_{0\le m<k}\|\nabla F({\boldsymbol{\theta}}_m)\|^2
    =\mathcal O(k^{-1}),\qquad \text{a.s.}$}
& \ref{thm:nonconvex_as} \\

Diminishing PS / ALS: almost sure convergence
& & & \\

Convex
& $\sigma^2<\infty$
& $F(\bar{\boldsymbol{\theta}}_k^{\rm w})-F({\boldsymbol{\theta}}^\star)
   =\mathcal O(S_k^{-1}),\qquad \text{a.s.}$
& \ref{thm:convex_as_dec} \\

Non-convex
& Weak growth
& $\min_{0\le m\le k}\|\nabla F({\boldsymbol{\theta}}_m)\|^2
   =\mathcal O(S_k^{-1}),\qquad \text{a.s.}$
& \ref{thm:nonconvex_as_dec} \\

\end{tblr}
\raggedright
\endgroup
\end{table}

%% file: Sections/2.background.tex
\section{Preliminaries}
\label{sec:background}

\subsection{Polyak step size and Armijo line search for SHB}

Polyak step size (PS) was originally proposed for subgradient methods~\citep{polyak1987introduction}.
PS determines the step size using the objective value and the norm of the subgradient, and enjoys particularly strong convergence guarantees for convex problems. 
Armijo line search (ALS), on the other hand, selects a step size satisfying a sufficient-decrease condition through a backtracking procedure~\citep{armijo1966minimization}.
ALS is mainly designed for differentiable objectives, which ensures that a step size satisfying the Armijo rule can be found.
Both PS and ALS have been successfully extended to SGD and shown to enjoy favorable convergence guarantees~\citep{sls, SPS}.
Throughout, we assume that $F$ is bounded below and $f(\cdot;x)$ is bounded below for $p_x$-almost every $x$. Let $\boldsymbol{\theta}^\star \in \arg\min_{\boldsymbol{\theta}} F(\boldsymbol{\theta})$ be a global minimizer of $F$, and denote $f^\star(x):=\inf_{\boldsymbol{\theta}} f(\boldsymbol{\theta};x)>-\infty$.

We now introduce the forms of PS and ALS for SHB considered in this paper, termed SHB-PS and SHB-ALS, respectively.
Following~\citet{MomSPS}, the form of SHB-PS is consistent with that of PS in SGD~\citep{SPS}.
On top of this form, we include an additional scaling factor depending on the momentum parameter $\beta$, leading to
\begin{align} 
\label{eq:shb-ps}
\gamma_{k,\mathrm{SHB-PS}} = \min\left\{ \frac{\left(1-\sqrt{\beta}\right)^2 \left(f(\theta_k;x_k)-f^\star(x_k)\right)} {2c\|\nabla f(\theta_k;x_k)\|^2}, \gamma_{\max} \right\}, 
\end{align}
where $c$ is a hyperparameter, and $\gamma_{\max}$ is a prescribed upper bound to prevent the step size from becoming unbounded.
In practice, $f^\star(x)$ can also be replaced by any valid
lower bound on $f(\cdot;x)$~\citep{sps_dec}.
Note that the resulting step size is equivalent to those in~\citep{MomSPS,SPS}, since the additional constant factor can always be absorbed into the hyperparameter $c$.

For SHB-ALS, we introduce a slightly modified Armijo rule as
\begin{align} 
\label{eq:shb-ls}
f\left(\tilde{\boldsymbol\theta}_{k+1};x_k\right) \le f(\boldsymbol\theta_k;x_k) - c\tilde{\gamma}_k \|\nabla f(\boldsymbol\theta_k;x_k)\|^2,
\end{align}
where $c\in(0,1)$, and
\begin{align}
\label{eq:shb-ls-parameters}
\tilde{\gamma}_k = \frac{2\gamma_k}{\left(1-\sqrt{\beta}\right)^2}, \qquad \tilde{\boldsymbol\theta}_{k+1} = \boldsymbol\theta_k - \tilde{\gamma}_k\nabla f(\boldsymbol\theta_k;x_k).
\end{align}
We define $\gamma_{k,\mathrm{SHB-ALS}}$ as the largest $\gamma_k$ satisfying the above condition, obtained by a backtracking search initialized at $\gamma_{\max}$ with decay factor $\omega\in(0,1)$.

From these definitions, we immediately obtain the following basic lemma.
\begin{lemma}
\label{lemma:basic_step_size}
For $\gamma_k=\gamma_{k,\mathrm{SHB-PS}}$ or $\gamma_k=\gamma_{k,\mathrm{SHB-ALS}}$, we have
\begin{align}
    \gamma_k
    \|\nabla f(\boldsymbol{\theta}_k;x_k)\|^2
    \le
    \frac{(1-\sqrt{\beta})^2}{2c}
    \left(
    f(\boldsymbol{\theta}_k;x_k)-f^\star(x_k)
    \right).
\end{align}
\end{lemma}
Lemma~\ref{lemma:basic_step_size} follows directly from the definitions of SHB-PS and SHB-ALS.
This unified formula allows us to analyze the two algorithms within the same framework in the sequel.

% 从这一形式出发，我们立刻得到如下的结论：
% \begin{lemma}
% \label{lemma:basic_step_size}
%     对于$\gamma_k=\gamma_{k, \mathrm{SHB-PS}}$或$\gamma_k=\gamma_{k, \mathrm{SHB-LS}}$，我们有
%     \begin{align}
%         \gamma_{k}\|\nabla f(\boldsymbol{\theta_k};x_k)\|^2\le\frac{(1-\sqrt{\beta})^2}{2c}\left(f(\boldsymbol{\theta_k};x_k)-f^\star(x_k)\right).
%     \end{align}
% \end{lemma}
% Lemma~\ref{lemma:basic_step_size}可以由SHB-PS和SHB-LS的定义中立刻得到。这一统一形式使得我们后面能够统一地分析两种算法。

% \subsection{Assumptions and basic bound for the step sizes}

% 在所有的分析中，我们要求目标函数是一致Lipschitz smooth的，as stated in the following assumption。
% \begin{assumption}
%     Assume $f(\boldsymbol{\theta}; x)$ has $L(x)$-Lipschitz gradients，i.e.,
%     \begin{align}
%         补充L-smooth的定义
%     \end{align}
%     and $L:=\supp L(x)\lt \infty$。
% \end{assumption}
% 凸目标的分析中假设每个子目标函数均是凸的如下
% \begin{assumption}
%     Assume $f(\boldsymbol{\theta}; x)$ is convex with respect to $\theta$ over $p_x$ almost everywhere，i.e.,
%     \begin{align}
%         补充convex的定义
%     \end{align}
% \end{assumption}
% 对于非凸目标中，我们做weak-growth condition，which是已知的最弱的非凸分析的假设之一，as stated as follows。
% \begin{assumption}
%     我们称问题(1)满足weak growth condition, if
%     \begin{align}
%         补充weak growth condition的定义。
%     \end{align}
% \end{assumption}

\subsection{Assumptions and basic bounds for the step sizes}

Throughout the analysis, we assume that the stochastic objective is uniformly Lipschitz smooth.
\begin{assumption}[Smoothness]
\label{assump:smoothness}
Assume that $f(\boldsymbol{\theta};x)$ has $L(x)$-Lipschitz continuous gradients with respect to $\boldsymbol{\theta}$ for $p_x$-almost every $x$, i.e.,
\begin{align}
    \|\nabla f(\boldsymbol{\theta};x)-\nabla f(\boldsymbol{\theta}';x)\|
    \le
    L(x)\|\boldsymbol{\theta}-\boldsymbol{\theta}'\|,
    \qquad
    \forall \boldsymbol{\theta},\boldsymbol{\theta}' .
\end{align}
Moreover, we assume that
\begin{align}
    L := \sup_{x\sim p_x} L(x) < \infty .
\end{align}
\end{assumption}
Under Assumption~\ref{assump:smoothness}, the step sizes of SHB-PS and SHB-ALS are uniformly bounded from below and above. The proof follows the standard arguments for SPS and stochastic line search~\citep{SPS,sls}.
\begin{lemma}[Bounds on the step sizes]
\label{lemma:step_size_bounds}
Suppose Assumption~\ref{assump:smoothness} holds. For SHB-PS, we have
\begin{align}
    \gamma_{\min,\mathrm{PS}}
    :=
    \min\left\{
    \frac{\left(1-\sqrt{\beta}\right)^2}{4cL},
    \gamma_{\max}
    \right\}
    \le
    \gamma_{k,\mathrm{SHB-PS}}
    \le
    \gamma_{\max}.
\end{align}
For SHB-ALS, we have
\begin{align}
    \gamma_{\min,\mathrm{LS}}
    :=
    \min\left\{
    \gamma_{\max},
    \left(1-\sqrt{\beta}\right)^2
    \frac{\omega(1-c)}{L}
    \right\}
    \le
    \gamma_{k,\mathrm{SHB-ALS}}
    \le
    \gamma_{\max}.
\end{align}
\end{lemma}
Below, we use $\gamma_{\min}$ to denote the lower bound on both step sizes in the unified analysis.

For the analysis of convex objectives, we assume that each stochastic component is convex.
\begin{assumption}[Convexity]
\label{assump:convexity}
Assume $f(\boldsymbol{\theta};x)$ is convex with respect to $\boldsymbol{\theta}$ for $p_x$-almost every $x$, i.e.,
\begin{align}
    f(\boldsymbol{\theta}';x)
    \ge
    f(\boldsymbol{\theta};x)
    +
    \left\langle
    \nabla f(\boldsymbol{\theta};x),
    \boldsymbol{\theta}'-\boldsymbol{\theta}
    \right\rangle,
    \qquad
    \forall \boldsymbol{\theta},\boldsymbol{\theta}' .
\end{align}
\end{assumption}
For the strongly convex analysis, we additionally use the following standard definition.
\begin{definition}[Strong convexity]
\label{def:strong_convexity}
A differentiable function $F$ is $\mu$-strongly convex if there exists $\mu>0$ such that, for all
$\boldsymbol{\theta},\boldsymbol{\theta}'$,
\begin{align}
    F(\boldsymbol{\theta}')
    \ge
    F(\boldsymbol{\theta})
    +
    \left\langle
        \nabla F(\boldsymbol{\theta}),
        \boldsymbol{\theta}'-\boldsymbol{\theta}
    \right\rangle
    +
    \frac{\mu}{2}
    \left\|
        \boldsymbol{\theta}'-\boldsymbol{\theta}
    \right\|^2.
\end{align}
\end{definition}

We also assume that the optimal difference is finite.
\begin{assumption}[Finite Optimal Difference]
\label{assump:finite_optimal_difference}
We assume that
\begin{align}
    \sigma^2
    :=
    \mathbb{E}_{x\sim p_x}
    \left[
    f(\boldsymbol{\theta}^\star;x)-f^\star(x)
    \right]
    <
    \infty .
\end{align}
\end{assumption}
The case $\sigma^2=0$ corresponds to the interpolation setting~\citep{sls}, where the global minimizer of $F$ also minimizes each stochastic objective almost surely.

For general non-convex objectives, we impose the weak growth condition~\citep{SPS}.
\begin{assumption}[Weak Growth Condition]
\label{assump:wgc}
We say Problem~\eqref{eq:stochastic_optimization_problem} satisfies the weak growth condition if there exist constants $\rho > 0$ and $\delta \ge 0$ such that, for all $\boldsymbol{\theta}$,
\begin{align}
    \mathbb{E}_{x\sim p_x}
    \left[
    \|\nabla f(\boldsymbol{\theta};x)\|^2
    \right]
    \le
    \rho
    \|\nabla F(\boldsymbol{\theta})\|^2
    +
    \delta .
\end{align}
\end{assumption}
When $\delta=0$, the weak growth condition reduces to the strong growth condition~\citep{schmidt2013fast}, which can be viewed as a non-convex analogue of interpolation.

% Then under Assumption~\ref{assump:smoothness}, SHB-PS and SHB-ALS are bounded as the following lemma, which可以follow经典的推导from\citep{SPS, sls}.
% \begin{lemma}
%     Assume Assumption~\ref{assump:smoothness} holds, then
%     $$
%     \gamma_{\min, \mathrm{PS}}=\min\left\{\frac{\left(1-\sqrt{\beta}\right)^2}{4cL},\gamma_\max\right\}\le\gamma_{k,\mathrm{SHB-PS}}\le\gamma_\max,
%     $$
%     and
%     $$
%     \gamma_{\min, \mathrm{LS}}=\min\left\{\gamma_\max,\left(1-\sqrt{\beta}\right)^2 \frac{\omega(1-c)}{L}\right\}\le\gamma_{k, \mathrm{SHB-ALS}}\le\gamma_{\max}.
%     $$
% \end{lemma}
% 下文在对两个步长的统一分析中，我们直接使用$\gamma_\min$来表示步长的下界。

%% file: Sections/3.general_convergence.tex
\section{General convergence with standard PS and ALS}
\label{sec:main_results}
\subsection{Decoupling lemmas for the SHB dynamics}

% SHB分析的一个难点是动量项中历史的梯度与当前迭代存在强耦合。我们handle这一问题通过直接对SHB迭代的分析，给出两个解耦引理分别针对convex和non-convex的分析。
% 首先，下面的Lemma给出了相邻迭代点之差和动量项的解耦形式。
% \begin{lemma}
% Denote $\theta_0, \theta_1, ...,\theta_k$ as the points obtained by the SHB~\eqref{eq:shb}, then the following holds
% \begin{align}
% &(\theta_k-\theta^\star)\sum_{m=0}^k\beta^{k-m}\gamma_m\nabla f(\theta_m; x_m)\nonumber\\
% =&\sum_{m=0}^k\beta^{k-m}\gamma_m(\theta_m-\theta^\star)^T\nabla f(\theta_m; x_m)-\sum_{n=0}^{k-1}\beta^{k-n}\left\|\sum_{m=0}^n\beta^{n-m}\gamma_m\nabla f(\theta_m; x_m)\right\|^2.
% \end{align}
% \end{lemma}
% Lemma~\ref{lemma:decouple_convex}解耦了当前迭代和历史的梯度。Also note that右手边的第一项自然与凸性相关，而第二项又能利用PS和ALS的基本性质（Lemma~\ref{lemma:basic_step_size}）。这使得我们能够实现凸函数的分析。

The main difficulty in analyzing SHB is that the historical gradients accumulated in the momentum term are strongly coupled with the current iterate.
We handle this issue by directly analyzing the SHB recursion and establishing two decoupling lemmas tailored to the convex and non-convex analyses, respectively. 
The following lemma decouples the current-iterate difference from the accumulated momentum term. 
\begin{lemma} 
\label{lemma:decouple_convex} 
Let $\boldsymbol{\theta}_0,\boldsymbol{\theta}_1,\ldots,\boldsymbol{\theta}_k$ be the iterates generated by SHB~\eqref{eq:shb}. 
Then, 
\begin{align} 
& \left\langle \boldsymbol{\theta}_k-\boldsymbol{\theta}^\star, \sum_{m=0}^k \beta^{k-m}\gamma_m \nabla f(\boldsymbol{\theta}_m;x_m) \right\rangle \nonumber\\
=& \sum_{m=0}^k \beta^{k-m}\gamma_m \left\langle \boldsymbol{\theta}_m-\boldsymbol{\theta}^\star, \nabla f(\boldsymbol{\theta}_m;x_m) \right\rangle - \sum_{n=0}^{k-1} \beta^{k-n} \left\| \sum_{m=0}^n \beta^{n-m}\gamma_m \nabla f(\boldsymbol{\theta}_m;x_m) \right\|^2 . 
\end{align} 
\end{lemma}
% Lemma~\ref{lemma:decouple_convex} decouples the current iterate from the historical gradients.
%.
We note that the first term on the right-hand side of Lemma~\ref{lemma:decouple_convex} is naturally connected to convexity, while the second term can be controlled using the basic property of PS and ALS in Lemma~\ref{lemma:basic_step_size}.
This decomposition therefore enables our analysis for convex objectives.

% 下面的定理则给出了相邻目标函数值满足的不等式。
% \begin{lemma}
%     Assume $\theta_0, \theta_1, ...,\theta_k$ are the points obtained by the Stochastic Heavy Ball algorithm, and under Assumption~\ref{assump:smoothness}, then
%     \begin{align}
%     F(\theta_{k+1})-F(\theta_k)\le -\sum_{m=0}^k\beta^{k-m}\gamma_m\left\langle\nabla F(\theta_m),\nabla f(\theta_m,x_m)\right\rangle+L\sum_{n=0}^k\beta^{k-n}\left\|\sum_{m=0}^n\beta^{n-m}\gamma_m\nabla f(\theta_m, x_m)\right\|^2.
%     \end{align}
% \end{lemma}

The next lemma provides an upper bound on the difference between successive objective values.
\begin{lemma}
\label{lemma:decouple_nonconvex}
Let $\boldsymbol{\theta}_0,\boldsymbol{\theta}_1,\ldots,\boldsymbol{\theta}_k$ be the iterates generated by SHB~\eqref{eq:shb}. 
Under Assumption~\ref{assump:smoothness}, we have
\begin{align}
    F(\boldsymbol{\theta}_{k+1})
    -
    F(\boldsymbol{\theta}_k)
    \le
    &
    -
    \sum_{m=0}^k
    \beta^{k-m}\gamma_m
    \left\langle
    \nabla F(\boldsymbol{\theta}_m),
    \nabla f(\boldsymbol{\theta}_m;x_m)
    \right\rangle
    \nonumber\\
    &
    +
    L
    \sum_{n=0}^k
    \beta^{k-n}
    \left\|
    \sum_{m=0}^n
    \beta^{n-m}\gamma_m
    \nabla f(\boldsymbol{\theta}_m;x_m)
    \right\|^2 .
\end{align}
\end{lemma}
Lemma~\ref{lemma:decouple_nonconvex} characterizes the relationship between the gradients of the objective function and the successive objective values along the SHB trajectory.
This connection allows us to translate the decrease in objective values into a best-iterate stationarity guarantee for general non-convex objectives.
The complete proofs of Lemma~\ref{lemma:decouple_convex} and Lemma~\ref{lemma:decouple_nonconvex} are deferred to Appendix~\ref{appendix:assumptions_and_lemmas}.

\subsection{Expected convergence rates for SHB-PS and SHB-ALS}
\label{sec:expected_convergence_rates}

% Now we are ready to present the general convergence rates for SHB-PS and SHB-ALS. 下面的定理给出了general 凸目标上的SHB-PS和SHB-ALS的收敛速率。
% \begin{theorem}
%     Assume $F(\theta)$ is convex and $f(\theta;x)$ is convex for all $x\in \mathcal{D}$. Then the convergence rate of SHB-LS or SHB-LS with $c>1/2$ is given as
%     \begin{align}
%     \mathbb{E}\left(F\left(\overline \theta_k\right)-F(\theta^\star)\right)\le \frac{A}{k}+M\cdot\frac{k+1}{k}\sigma^2,
%     \end{align}
%     where $\overline\theta_k=\frac{1}{k}\sum_{m=0}^{k-1}\theta_m$, and
%     \begin{align}
%         A=\frac{\left\|\theta_0-\theta^\star\right\|^2}{(1+\beta)\left(2-1/c\right)\gamma_{\min}},\quad
%         M=\frac{1}{1-\beta^2}\left(\frac{2\gamma_{\max}}{\gamma_{\min}(2-1/c)}-1\right).
%     \end{align}
% \end{theorem}
We are now ready to present the general convergence rates for SHB-PS and SHB-ALS.
First, we consider strongly convex objectives, for which the decoupling analysis yields a linear convergence guarantee in expectation for the last iterate.
\begin{theorem}
\label{thm:strongly_convex_general}
Under Assumptions~\ref{assump:smoothness}, \ref{assump:convexity}, and~\ref{assump:finite_optimal_difference}, suppose further that $F$ is $\mu$-strongly convex (Definition~\ref{def:strong_convexity}).
Then, for SHB-PS or SHB-ALS with $c\ge 1/2$, we have
\begin{align}
    \mathbb{E}
    \left\|
        \boldsymbol{\theta}_k-\boldsymbol{\theta}^\star
    \right\|^2
    \le
    B r^k
    +
    \frac{2\gamma_{\max}}{\mu\gamma_{\min}}\sigma^2,
\end{align}
where $B>0$ is a constant depending on the initialization, and
\begin{align}
    r
    =
    \frac{
        1-\mu\gamma_{\min}
        +
        \sqrt{
            (1-\mu\gamma_{\min})^2
            -4\mu\gamma_{\min}\beta
        }
    }{2}
    \in [0,1).
\end{align}
\end{theorem}
\paragraph{Proof sketch.}
Combining Lemma~\ref{lemma:basic_step_size} and Lemma~\ref{lemma:decouple_convex} with strong convexity yields a recursive inequality for the expected squared distance to $\boldsymbol{\theta}^\star$.
After subtracting the residual term, the resulting sequence can be controlled through a second-order recurrence induced by momentum, giving the stated linear rate.
The complete proof is deferred to Appendix~\ref{appendix:proof_thm_strongly_convex}.

Theorem~\ref{thm:strongly_convex_general} shows that SHB-PS and SHB-ALS converge linearly to a neighborhood of the optimum determined by the optimal difference $\sigma^2$.
In particular, under interpolation ($\sigma^2=0$), the last iterate converges linearly to the exact optimum in expectation.
We also note that when $\beta=0$, SHB reduces to SGD and the contraction factor reduces to $r=1-\mu\gamma_{\min}$, recovering the classical SPS rate $\mathcal{O}((1-\mu\gamma_{\min})^k)$ as shown in~\cite{SPS}.

The following theorem gives the rate of SHB-PS and SHB-ALS for general convex objectives.
\begin{theorem}
\label{thm:convex_general}
Under Assumptions~\ref{assump:smoothness}, \ref{assump:convexity}, and~\ref{assump:finite_optimal_difference}, for SHB-PS or SHB-ALS with $c>1/2$, we have
\begin{align}
    \mathbb{E}
    \left[
    F\left(\bar{\boldsymbol{\theta}}_k\right)
    -
    F(\boldsymbol{\theta}^\star)
    \right]
    \le
    \frac{A}{k}
    +
    M\cdot
    \frac{k+1}{k}
    \sigma^2,
\end{align}
where $\bar{\boldsymbol{\theta}}_k=\frac{1}{k}\sum_{m=0}^{k-1}\boldsymbol{\theta}_m$,
and
\begin{align}
    A=\frac{\left\|\boldsymbol{\theta}_0-\boldsymbol{\theta}^\star\right\|^2}
    {(1+\beta)\left(2-1/c\right)\gamma_{\min}},\quad
    M=\frac{1}{1-\beta^2}\left(\frac{2\gamma_{\max}}{\gamma_{\min}(2-1/c)}-1\right).
\end{align}
\end{theorem}
\paragraph{Proof sketch.}
Lemma~\ref{lemma:decouple_convex} together with Lemma~\ref{lemma:basic_step_size} controls the momentum terms and yields a recursion for the cumulative suboptimality.
Taking expectation, summing over iterations, and applying Jensen's inequality to $\bar{\boldsymbol{\theta}}_k$ gives the desired results.
The complete proof is deferred to Appendix~\ref{appendix:proof_thm_convex}.

% Theorem~\ref{thm:convex_general}说明SHB-PS和SHB-ALS在凸函数上的$O(1/k)$的速率to a neighborhood governed by the optimal difference。 For SHB-PS, this regime
% is close to MomSPS-style~\citep{MomSPS} convex expected guarantees. Note that MomSPS~\citep{MomSPS}需要对$\beta$的上限的限制. Our results对$\beta$没有任何限制，能够更好地覆盖实用场景中地大动量场景例如$\beta=0.9$。
Theorem~\ref{thm:convex_general} shows that SHB-PS and SHB-ALS achieve an $\mathcal{O}(1/k)$ rate for convex objectives up to a neighborhood governed by the optimal difference.
For SHB-PS, this result is closely related to the rate established for $\mathrm{MomSPS}_{\max}$~\citep{MomSPS}.
However, the analysis of $\mathrm{MomSPS}_{\max}$ requires an upper bound on the momentum parameter $\beta$.
%
% Our result imposes no restriction on $\beta$, and therefore better covers the large-momentum regimes often used in practice.
In comparison, our result does not impose an additional upper bound and thus applies to any fixed $\beta\in[0,1)$.
% Proofs are deferred to Appendix~\ref{appendix:proof_thm_convex}.

For general non-convex objectives satisfying the weak growth condition, the following theorem gives a best-iterate stationarity guarantee.
\begin{theorem}
\label{thm:nonconvex_general}
Under Assumptions~\ref{assump:smoothness} and~\ref{assump:wgc}, suppose that
$c\ge \rho/4$ for SHB-PS and $c>1-1/(\omega\rho)$ for SHB-ALS.
Let $\kappa=1/c$ for SHB-PS and $\kappa=4\omega(1-c)$ for SHB-ALS, and further assume that
\begin{align}
    0<\gamma_{\max}<
    \frac{(1-\sqrt{\beta})^2}{4\rho L}
    \left(
    1-\rho+
    \sqrt{(1-\rho)^2+2\rho(1+\rho)\kappa}
    \right).
\end{align}
Then, we have
\begin{align}
    \min_{0\le m\le k-1}
    \mathbb{E}\|\nabla F(\boldsymbol{\theta}_m)\|^2
    \le
    \frac{F(\boldsymbol{\theta}_0)-F(\boldsymbol{\theta}^\star)}
    {k(1+\beta)s_1}
    +
    \frac{k+1}{k}
    \frac{s_2\delta}{(1+\beta)s_1},
\end{align}
where
\begin{align}
    s_1=\frac{\gamma_{\max}+\gamma_{\min}}{2}-\rho q,
    \qquad
    s_2=\frac{q}{1-\beta},
    \qquad
    q=\frac{L\gamma_{\max}^2}{(1-\sqrt{\beta})^2}+\frac{\gamma_{\max}-\gamma_{\min}}{2}.
\end{align}
\end{theorem}
\paragraph{Proof sketch.}
Combining Lemma~\ref{lemma:decouple_nonconvex} with the Cauchy--Schwarz inequality yields a descent bound in terms of a weighted sum of expected gradient norms.
Summing this bound and taking the minimum gives the stated best-iterate convergence rate. The complete proof is deferred to Appendix~\ref{appendix:proof_thm_non_convex}.

Theorem~\ref{thm:nonconvex_general} shows that the best iterate of SHB-PS and SHB-ALS approaches a neighborhood of stationarity at an $\mathcal{O}(1/k)$ rate.
The hyperparameter conditions in Theorem~\ref{thm:nonconvex_general} ensure that the constants $s_1$ and $s_2$ are positive, and the detailed verification is also deferred to Appendix~\ref{appendix:proof_thm_non_convex}.
Specifically, the rate of SHB-PS matches the order of the non-convex guarantee of SPS~\citep{SPS}.
This is also consistent with existing theoretical observations that momentum does not necessarily improve the theoretical order of convergence rates~\citep{ganesh2023does, SGDM-APS}.

\subsection{Almost sure convergence under interpolation and strong growth}

Under interpolation or strong growth, stochastic optimization algorithms often enjoy stronger convergence guarantees~\citep{ma2018power, vaswani2019fast, sls, berrada2020training}.
In particular, exact convergence to the optimum or to stationarity can be obtained without diminishing step sizes, as the residual neighborhood terms in Theorems~\ref{thm:strongly_convex_general}, ~\ref{thm:convex_general} and~\ref{thm:nonconvex_general} vanish.
Moreover, we can further strengthen the convergence-in-expectation results to almost sure convergence.
The convex and non-convex cases are presented in the following two theorems, respectively.

\begin{theorem}
\label{thm:convex_as}
Assume that interpolation holds, i.e., $\sigma^2=0$, and that the conditions in Theorem~\ref{thm:convex_general} hold.
Then, for SHB-PS or SHB-ALS, we have
\begin{align}
    F(\boldsymbol{\theta}_k)
    \to
    F(\boldsymbol{\theta}^\star)
    \quad \text{a.s.},
    \qquad \text{and} \qquad
    F(\bar{\boldsymbol{\theta}}_k)
    -
    F(\boldsymbol{\theta}^\star)
    =
    \mathcal{O}(1/k)
    \quad \text{a.s.}
\end{align}
\end{theorem}

\begin{theorem}
\label{thm:nonconvex_as}
Assume that the strong growth holds, i.e., $\delta=0$, and that the conditions in Theorem~\ref{thm:nonconvex_general} hold.
Then, for SHB-PS or SHB-ALS, we have
\begin{align}
    \|\nabla F(\boldsymbol{\theta}_k)\|
    \to
    0
    \quad \text{a.s.},
    \qquad \text{and} \qquad
    \min_{0\le m\le k-1}
    \|\nabla F(\boldsymbol{\theta}_m)\|^2
    =
    \mathcal{O}(1/k)
    \quad \text{a.s.}
\end{align}
\end{theorem}
The proofs of Theorems~\ref{thm:convex_as} and~\ref{thm:nonconvex_as} essentially arise as byproducts of the analysis in the general setting.
Under interpolation or strong growth, the corresponding residual terms vanish, allowing us to show that a nonnegative sequence is summable and thereby establish almost sure convergence.
The detailed proofs are deferred to Appendices~\ref{appendix:proof_thm_as_convex} and~\ref{appendix:proof_thm_as_non_convex}.
Notably, under interpolation or strong growth, both SHB-PS and SHB-ALS also guarantee last-iterate convergence, which is particularly relevant in practice since the final iterate is typically used as the output of the optimization procedure.

% Under interpolation，随机优化算法具有更好的收敛速率结论。例如，无需danimishing步长即可收敛到精确最优点 / stationary point，which可以直接由Theroem~\ref{thm:convex_general}和\ref{thm:nonconvex_general}推出。Moreover，我们实际上能够得到更强的almost sure convergence results beyond convergence in expectation。凸和非凸的情形分别展示在下面的两个定理中。

% \begin{theorem}
%     Assume the interpolation holds, i.e., $\sigma^2=0$, and the conditions in Theorem~\ref{thm:convex_general} hold. Then we have 
%     $$
%     F(\theta_k)\rightarrow F(\theta^\star), \quad a.s.,
%     $$
%     and
%     $$
%     F\left(\overline \theta_k\right)-F(\theta^\star)=\mathcal{O}({1}/{k}), \quad a.s.
%     $$
% \end{theorem}
% \begin{theorem}
%     Assume $\delta=0$, and the conditions in Theorem~\ref{thm:nonconvex_general} hold. Then we have
% $$
% \|\nabla F(\theta_m)\|\rightarrow 0,\quad a.s.,
% $$
% and
% $$
% \min_{m=0,\dots, k-1}\|\nabla F(\theta_m)\|^2=\mathcal{O}(1/k),\quad a.s.
% $$
% \end{theorem}

% Theorem~\ref{thm:convex_as}和~\ref{thm:nonconvex_as}的证明实际上基本来自于general场景分析中的byproduct。在interpolation场景下，我们能够直接得到一个正序列的集级数和有限，thus能够得到almost sure的收敛。详细的推导见~\ref{appendix:proof_thm_as_convex}and ~\ref{appendix:proof_thm_as_non_convex}。Note that 在interpolation下，SHB-LS和SHB-PS同时也保证了last iteration的收敛。这更加贴近实践因为大部分场景中人们都会使用last iteration的迭代结果。

%% file: Sections/4.exact_convergence_beyond.tex
\section{Exact convergence with diminishing step sizes}
\label{sec:diminishing}

The preceding results establish convergence guarantees for SHB-PS and SHB-ALS under general stochastic objectives. In the non-interpolation regime, however, the residual stochasticity at the optimum generally leads to convergence only to a neighborhood of the optimum or a stationary point. To achieve exact asymptotic convergence, we therefore consider variants of SHB-PS and SHB-ALS following the diminishing step sizes in~\citet{sebbouh2021almost}.

Specifically, we consider the stochastic heavy-ball iteration with a time-varying momentum,
\begin{align}
\boldsymbol{\theta}_{k+1}
=
\boldsymbol{\theta}_k
-\gamma_k\nabla f(\boldsymbol{\theta}_k;x_k)
+\beta_k(\boldsymbol{\theta}_k-\boldsymbol{\theta}_{k-1}),
\end{align}
where $\beta_k\in[0,1)$ denotes the momentum parameter at iteration $k$.
The diminishing variant of SHB-PS is defined as
\begin{align}
    \gamma_{k,\mathrm{SHB\text{-}PS}_{\mathrm{dec}}}=
    \eta_k
    \min {\left\{
    \frac{f(\boldsymbol{\theta}_k;x_k)-f^*(x_k)}
    {c\|\nabla f(\boldsymbol{\theta}_k;x_k)\|^2},
    \gamma_{\max,k}
    \right\}},
\end{align}
where both $\eta_k$ and $\gamma_{\max,k}$ are iteration-dependent sequences.
% \cite{sebbouh2021almost}
Similarly, let
$\gamma_{k,\mathrm{SHB\text{-}ALS}}(\gamma_{\max,k})$
denote the largest $\gamma_k$ obtained by backtracking from $\gamma_{\max,k}$ with backtracking factor $\omega$, such that
\begin{align}
f\left(\tilde{\boldsymbol{\theta}}_{k+1};x_k\right)
\le
f(\boldsymbol{\theta}_k;x_k)
-c\gamma_{k}
\|\nabla f(\boldsymbol{\theta}_k;x_k)\|^2,\quad \tilde{\boldsymbol{\theta}}_{k+1}=\boldsymbol{\theta}_k-\gamma_k\nabla f(\boldsymbol{\theta}_k;x_k).
\end{align}
The diminishing variant of SHB-ALS is then defined as
\begin{align}
\gamma_{k,\mathrm{SHB\text{-}ALS}_{\mathrm{dec}}}=
\eta_k
\gamma_{k,\mathrm{SHB\text{-}ALS}}(\gamma_{\max,k}).
\end{align}
The following lemma provides lower and upper bounds for the diminishing step-size rules, mirroring Lemma~\ref{lemma:step_size_bounds} for the non-diminishing case.
\begin{lemma}
\label{lemma:dec_stepsize_bounds}
Under Assumption~\ref{assump:smoothness}, we have
\begin{align}
\eta_k
\min\left\{
\frac{1}{2cL},
\gamma_{\max,k}
\right\}
\le
\gamma_{k,\mathrm{SHB\text{-}PS}_{\mathrm{dec}}}
\le
\eta_k\gamma_{\max,k},
\end{align}
and
\begin{align}
\eta_k
\min\left\{
\gamma_{\max,k},
\frac{2\omega(1-c)}{L}
\right\}
\le
\gamma_{k,\mathrm{SHB\text{-}ALS}_{\mathrm{dec}}}
\le
\eta_k\gamma_{\max,k}.
\end{align}
\end{lemma}
Similarly, below we use the unified bounds for both step sizes as
\begin{align}
\eta_k\gamma_{\min,k}
\le
\gamma_k
\le
\eta_k\gamma_{\max,k}.
\label{eq:dec_stepsize_bound}
\end{align}
Throughout the analysis for diminishing step sizes, we impose the following conditions on the momentum and step-size schedules.
\begin{assumption}
\label{assump:as_decay}
We assume
\begin{enumerate}
    \item[(i)] $\{\beta_k\}_{k\ge1}$ is non-increasing,
    $\beta_k\in[0,1)$, and $\sum_{k=1}^{\infty}\beta_k < \infty$.
    
    \item[(ii)] $\{\eta_k\gamma_{\max,k}\}_{k\ge1}$ is non-increasing, and $\lim_{k\to\infty}\gamma_{\max,k}=0$.
    
    \item[(iii)] $
        \sum_{k=1}^{\infty}\eta_k\gamma_{\max,k}
        = \infty,
        \text{ and }
        \sum_{k=1}^{\infty}\eta_k^2\gamma_{\max,k}
        < \infty.
    $
\end{enumerate}
\end{assumption}
We note that the summability condition on $\{\beta_k\}$ in (i) has also
been adopted in previous convergence analyses of momentum methods
~\citep{sun2019non, sun2020nonergodic, rosasco2015stochastic, j2018on, luo2018adaptive}. Decay conditions on
$\gamma_{\max,k}$ in (ii), together with the step-size conditions in
(iii), are also used in the analysis of SHB in~\cite{sebbouh2021almost}. These conditions are closely related to the Robbins--Monro condition in stochastic approximation methods~\citep{robbins1951stochastic}.

We are now ready to present the results of $\text{SHB-PS}_\text{dec}$ and $\text{SHB-ALS}_\text{dec}$. First, we establish convergence to the optimal objective value
for a weighted average of the iterates in the convex setting.
\begin{theorem}
\label{thm:convex_as_dec}
Suppose Assumptions~\ref{assump:smoothness},
\ref{assump:convexity}, \ref{assump:finite_optimal_difference},
and~\ref{assump:as_decay} hold.
Assume further that there exists a sufficiently large integer $K$ such that
\begin{align}
\label{eq:condition_c_as_convex}
    c \ge
    \sup_{k>K}
    \frac{\eta_k}{
        1+\beta_k
        -\frac{\beta_k}{1+\beta_k}
        \frac{\gamma_{\max,k}}{\gamma_{\min,k}}
    }.
\end{align}
Then, for either $\text{SHB\text{-}PS}_{\text{dec}}$ or
$\text{SHB\text{-}ALS}_{\text{dec}}$, we have
\begin{align}
    F(\bar{\boldsymbol{\theta}}_k^{\rm w})
    -F(\boldsymbol{\theta}^\star)
    =
    \mathcal{O}\left(
        \frac{1}{\sum_{m=1}^k \eta_m\gamma_{\max,m}}
    \right)
    \quad \text{a.s.},
\end{align}
where
\begin{align}
    \bar{\boldsymbol{\theta}}_k^{\rm w}
    :=
    \frac{
        \sum_{m=1}^k
        \eta_m\gamma_{\min,m}\boldsymbol{\theta}_m
    }{
        \sum_{m=1}^k \eta_m\gamma_{\min,m}
    }.
\end{align}
\end{theorem}
\paragraph{Proof sketch.}
We first derive recursive bounds for the squared distance to an optimum
and the squared difference between successive iterates.
Applying the Robbins--Siegmund theorem twice establishes almost sure summability of the squared iterate differences and weighted objective gaps. Applying Jensen's inequality gives the stated rate.
The complete proof is deferred to Appendix~\ref{appendix:proof_thm_as_convex_dec}.

Here we give a quick remark on the condition on $c$ in~\eqref{eq:condition_c_as_convex}.
Under Assumption~\ref{assump:as_decay}, the denominator in~\eqref{eq:condition_c_as_convex} converges to $1$.
Consequently, if $\sup_{k>K}\eta_k<1$ for some $K$, we can always choose a $c\in(0,1)$ satisfying~\eqref{eq:condition_c_as_convex}.
This is particularly relevant for $\mathrm{SHB\text{-}ALS}_{\mathrm{dec}}$,
which requires $c<1$ to ensure that backtracking finds a step size satisfying the Armijo rule. 

The following theorem then establishes the convergence rate
for the best-iterate squared gradient norm in the non-convex setting.
\begin{theorem}
\label{thm:nonconvex_as_dec}
Suppose Assumptions~\ref{assump:smoothness},
\ref{assump:wgc}, and~\ref{assump:as_decay} hold.
For either $\text{SHB\text{-}PS}_{\text{dec}}$ or
$\text{SHB\text{-}ALS}_{\text{dec}}$, we have
\begin{align}
    \min_{1\le m\le k}
    \|\nabla F(\boldsymbol{\theta}_m)\|^2
    =
    \mathcal{O}\left(
        \frac{1}{\sum_{m=1}^k \eta_m\gamma_{\max,m}}
    \right)
    \quad \text{a.s.}
\end{align}
\end{theorem}
\paragraph{Proof sketch.}
Using smoothness and the weak growth condition, we derive a Lyapunov
recursion combining the objective gap with a weighted difference
between successive iterates.
The assumptions ensure that the error terms are summable.
Applying the Robbins--Siegmund theorem and a weighted minimum argument yields the stated rate.
The complete proof is deferred to Appendix~\ref{appendix:proof_thm_as_nonconvex_dec}.

For the decaying variants of SHB-PS and SHB-ALS, Theorem~\ref{thm:convex_as_dec} and~\ref{thm:nonconvex_as_dec} both establish almost sure convergence.
\citet{MomSPS} also considers decaying variants of SPS, namely MomDecSPS and MomAdaSPS, which likewise guarantee convergence to the exact solution, but only in expectation.
We would like to note that the $\mathcal{O}$ rates in Theorem~\ref{thm:convex_as_dec} and~\ref{thm:nonconvex_as_dec} can be further strengthened to $o$ rates by exploiting the tail property and smoothness. We provide a more detailed proof of the $o$-refinement in Appendices~\ref{appendix:convex_little_o}
and~\ref{appendix:nonconvex_little_o}.

% 我们快速给一个remark on the condition on c in \eqref{eq:condition_c_as_convex}。由Assumption~\ref{assump:as_decay}，\eqref{eq:condition_c_as_convex}的分母会收敛到1。因此，只要存在一个K使得$\sup_{k>K} \eta_k<1$，我们就总能找到一个$c$满足条件。这对于$\text{SHB-ALS}_\text{dec}$很重要，因为它需要$c<1$以保证能够搜索到一个满足Armijo rule的步长。

% We note that the summable assumption for $\beta_k$ (in (i)) has also been 使用在previous的带动量的收敛性分析中~\cite{}。$\gamma_{\max, k}$的收敛性 (in (ii))和步长条件((iii))也是~\cite{sebbouh2021almost}中使用的，这也是随机梯度算法中证明收敛的标准条件~\cite{}。

% The first condition in Assumption~\ref{assump:as_decay} corresponds to a diminishing and summable momentum schedule. The remaining conditions are standard diminishing-step requirements: the cumulative effective step size remains unbounded, whereas the higher-order stochastic error terms are summable. Since $\gamma_{\min,k}=\gamma_{\max,k}$ for all sufficiently large $k$, the same asymptotic conditions can equivalently be expressed in terms of $\gamma_{\max,k}$ up to a finite number of initial iterations.

%% file: Sections/5.numerical_results.tex
\section{Numerical validation}
\label{sec:numerical_theory_aligned}

We examine whether the convergence behavior predicted by our
analysis is observed on controlled finite-sum problems satisfying
the theoretical assumptions.
In all experiments, $\boldsymbol{\theta}\in\mathbb{R}^{20}$ and
$p_x$ is the uniform distribution on $\{1,\ldots,n\}$, with $n=512$.
For convex objectives, we consider
\begin{align}
    f(\boldsymbol{\theta};i)
    =\frac12
    \bigl(\mathbf{a}_i^\top\boldsymbol{\theta}-b_i\bigr)^2,
    \qquad
    b_i
    =\mathbf{a}_i^\top\boldsymbol{\theta}^\star
      +\nu\epsilon_i.
    \label{eq:experiment_convex}
\end{align}
% where $\epsilon$ is a noise vector.
The matrix $\mathbf{A}$ with rows $\mathbf{a}_i^\top$ has full column rank, and the vector $\boldsymbol{\epsilon}$ with elements $\epsilon_i$ satisfies $\mathbf{A}^\top\boldsymbol{\epsilon}=\mathbf{0}$ and $\|\boldsymbol{\epsilon}\|^2/n=1$.
Thus, $\boldsymbol{\theta}^\star$ is the unique minimizer
of $F$.
We use $\nu=0$ as interpolation and $\nu=0.3$ as non-interpolation.
We draw $\boldsymbol{\theta}^{\star}\sim
\mathcal{N}(\mathbf{0},\mathbf{I}_{20})$ once and keep it fixed
across all runs.

For non-convex experiments, we use
\begin{align}
    f(\boldsymbol{\theta};i)
    =s_i\sum_{j=1}^{20}
    \left[\frac12\theta_j^2+2(1-\cos\theta_j)\right]
    -\boldsymbol{\xi}_i^\top\boldsymbol{\theta}.
\end{align}
Under strong growth, we set $\boldsymbol{\xi}_i=\mathbf{0}$
and assign $s_i=0.5$ and $s_i=1.5$ for $256$ components, respectively.
Beyond interpolation, we set $s_i=1$ and use perturbations
$\boldsymbol{\xi}_i\in\{-0.3,0.3\}^{20}$ occurring in opposite pairs.
$\boldsymbol{\theta}^{\star}$ is set to $\mathbf{0}$.
The two settings satisfy Assumption~\ref{assump:wgc} with
$(\rho,\delta)=(1.25,0)$ and $(1,1.8)$, respectively.
We initialize
$\boldsymbol{\theta}_{-1}=\boldsymbol{\theta}_0=\pi\mathbf{1}$.

\begin{figure}[t]
    \centering
    \includegraphics[width=\textwidth]{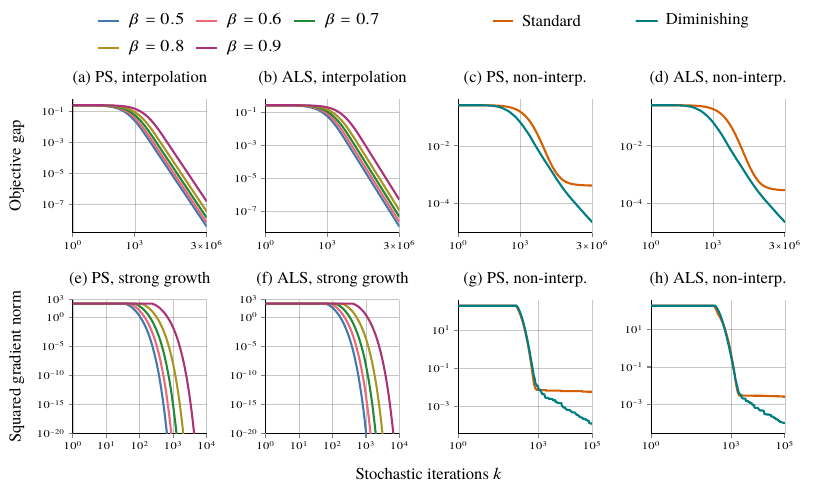}
    \caption{
Convergence of SHB with PS and ALS on convex (top) and non-convex (bottom) objectives.
Panels (a)--(b) and (e)--(f) show the standard, non-diminishing variants, while
(c)--(d) and (g)--(h) additionally include the diminishing variants.
For the standard step sizes, we report the empirical counterparts of the expected quantities in
Theorems~\ref{thm:convex_general} and~\ref{thm:nonconvex_general}:
the objective gap at the uniformly averaged iterate for convex objectives, and the running minimum of the squared full-gradient norm for non-convex objectives.
For the diminishing step sizes, which admit almost-sure convergence guarantees, we report the pointwise \textit{maximum} as a proxy across runs of the corresponding pathwise quantities:
the weighted-average objective gap for convex objectives and the best-iterate squared gradient norm for non-convex objectives, following
Theorems~\ref{thm:convex_as_dec} and~\ref{thm:nonconvex_as_dec}, respectively. 
    }
    \label{fig:main_text_theory_aligned}
\end{figure}
We implement PS and ALS as defined in
Sections~\ref{sec:background} and~\ref{sec:diminishing},
with $c=0.75$ and $\omega=0.5$.
For the standard (non-diminishing) variants, we use $\gamma_{\max}=1$
in the convex setting and $0.8$ times the upper bound in
Theorem~\ref{thm:nonconvex_general} in the non-convex setting.
We compare $\beta\in\{0.5,0.6,0.7,0.8,0.9\}$ under
interpolation or strong growth, and use $\beta=0.9$ otherwise.
For the diminishing variants, we set
\begin{align}
    \eta_k=0.5(1+k/10^4)^{-0.4},\quad
    \beta_k=0.9(1+k/10^4)^{-1.1},\quad
    \gamma_{\max,k}
    =\gamma_{\max,0}(1+k/10^4)^{-0.3}.
    \nonumber
\end{align}
We set $\gamma_{\max, 0}=0.02$ and $\gamma_{\max, 0}=2\gamma_{\max}^{\mathrm{std}}$ for convex and non-convex objectives, respectively. Here $\gamma_{\max}^{\mathrm{std}}$ denotes the upper bound used by the corresponding standard step sizes at $\beta=0.9$.
% For convex objectives, $\gamma_{\max,0}\!=\!0.02$. 
% For non-convex objectives, we set $\gamma_{\max,0}\!=\!2\gamma_{\max}^{\mathrm{std}}$,
% where $\gamma_{\max}^{\mathrm{std}}$ denotes the upper bound used by the corresponding standard step sizes at $\beta=0.9$.
% For non-convex objectives, we set $\gamma_{\max, 0}$ to $2$ times the upper bound of non-interpolation step sizes with $\beta=0.9$.
%
All experiments use 20 random seeds, and the metrics are chosen to match the corresponding theorems.

Figure~\ref{fig:main_text_theory_aligned} illustrates the behaviour of adaptive step sizes under different assumptions.
(a)--(b) and (e)--(f) validate the general convergence of non-diminishing SHB-PS and SHB-ALS under different constant $\beta$, which is consistent with Theorem~\ref{thm:convex_general}, \ref{thm:nonconvex_general} and the discussion in Section~\ref{sec:expected_convergence_rates}. Beyond interpolation, (c)--(d) and (g)--(h) show that non-diminishing step sizes converge only to a neighborhood, while the diminishing variants continuously converge, which is consistent with Theorem~\ref{thm:convex_as_dec} and \ref{thm:nonconvex_as_dec}.
More numerical results in machine learning tasks can be found in Appendix~\ref{appendix:numerical}.

%% file: Sections/6.conclusion.tex
\section{Conclusion}
\label{sec:conclusion}

We develop a unified convergence analysis for stochastic heavy ball methods with the Polyak step size and Armijo line search.
Building on a modified Armijo rule and two decoupling lemmas that isolate the effect of the momentum term, we establish expected convergence guarantees for both convex and non-convex objectives without imposing restrictive conditions on the momentum parameter.
Under interpolation or strong growth, our analysis further yields almost sure rates and last-iterate convergence.
Beyond the interpolation regime, we show that diminishing variants of~both step sizes exhibit almost-sure exact convergence.
% converge almost surely to the exact solution. 
%
Together, these results provide a more comprehensive theoretical understanding of Polyak step size and Armijo line search in stochastic heavy ball.

%
% A promising direction for future work is to develop variants of SHB-PS and SHB-ALS that attain exact convergence to the optimum or to stationarity beyond the interpolation regime, potentially through diminishing step-size schedules and adaptive momentum mechanisms.

\paragraph{Limitations and future work.}
Our analysis of the diminishing variants of SHB-PS and SHB-ALS still requires the momentum parameters ${\beta_k}$ to be summable.
This excludes commonly used constant momentum choices in practice, such as $\beta_k=0.9$.
Developing a more refined analysis that can accommodate non-summable or constant momentum therefore remains an important direction toward further narrowing the gap between theory and practice. 
Furthermore, since optimization algorithms are typically evaluated using their final iterates, extending the results to establish last-iterate convergence in the general setting would also be of significant interest.

%% file: Appendix/a.proof_of_lemmas.tex
\section{Proof of technical lemmas}
\label{appendix:assumptions_and_lemmas}

We first provide the proofs of all lemmas used in our analysis.
The proofs of Lemmas~\ref{lemma:basic_step_size} and~\ref{lemma:step_size_bounds} follow directly from the definitions or standard arguments in prior work~\citep{sls, SPS, ALR-SMAG, MomSPS, SGDM-APS}. We include them here for completeness.

\subsection{Proof of Lemma~\ref{lemma:basic_step_size}}

For SHB-PS, it follows directly from~\eqref{eq:shb-ps} that
\begin{align}
    \gamma_{k,\mathrm{SHB\text{-}PS}}
    \le
    \frac{
        (1-\sqrt{\beta})^2
        \left(f(\boldsymbol{\theta}_k;x_k)-f^\star(x_k)\right)
    }{
        2c\left\|\nabla f(\boldsymbol{\theta}_k;x_k)\right\|^2
    }.
\end{align}
Rearranging the above inequality gives the desired result.

For SHB-ALS, by~\eqref{eq:shb-ls} and~\eqref{eq:shb-ls-parameters}, any accepted step size satisfies
\begin{align}
    \frac{2c\gamma_{k,\mathrm{SHB\text{-}ALS}}}
    {(1-\sqrt{\beta})^2}
    \left\|\nabla f(\boldsymbol{\theta}_k;x_k)\right\|^2
    \le
    f(\boldsymbol{\theta}_k;x_k)
    -
    f(\tilde{\boldsymbol{\theta}}_{k+1};x_k)
    \nonumber\le
    f(\boldsymbol{\theta}_k;x_k)-f^\star(x_k).
\end{align}
The claimed bound follows immediately.

\subsection{Proof of Lemma~\ref{lemma:step_size_bounds}}

Under Assumption~\ref{assump:smoothness}, the descent lemma~\citep{nesterov2018lectures} gives
\begin{align}
\label{eq:descent_lemma}
    f(\boldsymbol{\theta}';x)
    \le\;&
    f(\boldsymbol{\theta};x)
    +
    \left\langle
        \nabla f(\boldsymbol{\theta};x),
        \boldsymbol{\theta}'-\boldsymbol{\theta}
    \right\rangle
    +
    \frac{L}{2}
    \left\|
        \boldsymbol{\theta}'-\boldsymbol{\theta}
    \right\|^2 .
\end{align}
It further implies~\citep{SPS}
\begin{align}
    f(\boldsymbol{\theta}_k;x_k)-f^\star(x_k)
    \ge
    \frac{1}{2L}
    \left\|\nabla f(\boldsymbol{\theta}_k;x_k)\right\|^2.
\end{align}
Substituting this inequality into the definition of SHB-PS yields
\begin{align}
    \gamma_{k,\mathrm{SHB\text{-}PS}}
    \ge
    \min\left\{
        \frac{(1-\sqrt{\beta})^2}{4cL},
        \gamma_{\max}
    \right\}
    =
    \gamma_{\min,\mathrm{PS}}.
\end{align}
% Together with the definition
% $\gamma_{k,\mathrm{SHB\text{-}PS}}\le\gamma_{\max}$,
% this proves the bound for SHB-PS.

For SHB-ALS, we first show that any step size satisfying
\begin{align}
\label{eq:shb-ls-condition}
    \gamma_k
    \le
    (1-\sqrt{\beta})^2\frac{1-c}{L}
\end{align}
satisfies the modified Armijo rule~\eqref{eq:shb-ls}.
By~\eqref{eq:descent_lemma}, we have
\begin{align}
\label{eq:proof_shb_ls_basic}
    &
    f(\boldsymbol{\theta}_k;x_k)
    -
    f\left(
        \boldsymbol{\theta}_k
        -
        \tilde{\gamma}_k
        \nabla f(\boldsymbol{\theta}_k;x_k);
        x_k
    \right)
    \nonumber
    \ge
    \tilde{\gamma}_k
    \left(
        1-\frac{L\tilde{\gamma}_k}{2}
    \right)
    \left\|\nabla f(\boldsymbol{\theta}_k;x_k)\right\|^2.
\end{align}
Since
$\tilde{\gamma}_k
=2\gamma_k/(1-\sqrt{\beta})^2$,
the above condition \eqref{eq:shb-ls-condition} on $\gamma_k$ ensures
$1-L\tilde{\gamma}_k/2\ge c$,
and hence~\eqref{eq:shb-ls} holds.
Accounting for the worst-case reduction by the backtracking factor $\omega$ therefore gives
\begin{align}
    \gamma_{k,\mathrm{SHB\text{-}ALS}}
    \ge
    \min\left\{
    \gamma_{\max},
    (1-\sqrt{\beta})^2\frac{\omega(1-c)}{L}
    \right\},
\end{align}
which completes the proof.

\subsection{Proof of Lemma~\ref{lemma:decouple_convex}}

Recall that the SHB recursion, with the initialization
$\boldsymbol{\theta}_{-1}=\boldsymbol{\theta}_0$, implies
\begin{align}
    \boldsymbol{\theta}_k
    =
    \boldsymbol{\theta}_{k-1}
    -
    \sum_{m=0}^{k-1}
    \beta^{k-1-m}\gamma_m
    \nabla f(\boldsymbol{\theta}_m;x_m).
\end{align}
Therefore,
\begin{align}
    &
    \left\langle
    \boldsymbol{\theta}_k-\boldsymbol{\theta}^\star,
    \sum_{m=0}^k
    \beta^{k-m}\gamma_m
    \nabla f(\boldsymbol{\theta}_m;x_m)
    \right\rangle 
    -
    \beta
    \left\langle
    \boldsymbol{\theta}_{k-1}-\boldsymbol{\theta}^\star,
    \sum_{m=0}^{k-1}
    \beta^{k-1-m}\gamma_m
    \nabla f(\boldsymbol{\theta}_m;x_m)
    \right\rangle
    \nonumber\\
    =&\,
    \gamma_k
    \left\langle
    \boldsymbol{\theta}_k-\boldsymbol{\theta}^\star,
    \nabla f(\boldsymbol{\theta}_k;x_k)
    \right\rangle
    -
    \beta
    \left\|
    \sum_{m=0}^{k-1}
    \beta^{k-1-m}\gamma_m
    \nabla f(\boldsymbol{\theta}_m;x_m)
    \right\|^2.
\end{align}
Recursively applying the above identity yields
\begin{align}
    &
    \left\langle
    \boldsymbol{\theta}_k-\boldsymbol{\theta}^\star,
    \sum_{m=0}^k
    \beta^{k-m}\gamma_m
    \nabla f(\boldsymbol{\theta}_m;x_m)
    \right\rangle
    \nonumber\\
    =&
    \sum_{m=0}^k
    \beta^{k-m}\gamma_m
    \left\langle
    \boldsymbol{\theta}_m-\boldsymbol{\theta}^\star,
    \nabla f(\boldsymbol{\theta}_m;x_m)
    \right\rangle
    -
    \sum_{n=0}^{k-1}
    \beta^{k-n}
    \left\|
    \sum_{m=0}^n
    \beta^{n-m}\gamma_m
    \nabla f(\boldsymbol{\theta}_m;x_m)
    \right\|^2,
\end{align}
which proves the lemma.

\subsection{Proof of Lemma~\ref{lemma:decouple_nonconvex}}

Under Assumption~\ref{assump:smoothness}, $F$ is also $L$-smooth. Therefore,
\begin{align}
\label{eq:first_inequality}
    F(\boldsymbol{\theta}_{k+1})&-F(\boldsymbol{\theta}_k)   
    \le
    \left\langle
    \nabla F(\boldsymbol{\theta}_k),
    \boldsymbol{\theta}_{k+1}-\boldsymbol{\theta}_k
    \right\rangle
    +
    \frac{L}{2}
    \left\|
    \boldsymbol{\theta}_{k+1}-\boldsymbol{\theta}_k
    \right\|^2
    \nonumber\\
    &=
    -\sum_{m=0}^k
    \beta^{k-m}\gamma_m
    \left\langle
    \nabla F(\boldsymbol{\theta}_k),
    \nabla f(\boldsymbol{\theta}_m;x_m)
    \right\rangle
    +
    \frac{L}{2}
    \left\|
    \sum_{m=0}^k
    \beta^{k-m}\gamma_m
    \nabla f(\boldsymbol{\theta}_m;x_m)
    \right\|^2 .
\end{align}
Similarly,
\begin{align}
\label{eq:second_inequality}
    &F(\boldsymbol{\theta}_{k-1})-F(\boldsymbol{\theta}_k)
        \nonumber\\
    \le&
    \sum_{m=0}^{k-1}
    \beta^{k-1-m}\gamma_m
    \left\langle
    \nabla F(\boldsymbol{\theta}_k),
    \nabla f(\boldsymbol{\theta}_m;x_m)
    \right\rangle
    +
    \frac{L}{2}
    \left\|
    \sum_{m=0}^{k-1}
    \beta^{k-1-m}\gamma_m
    \nabla f(\boldsymbol{\theta}_m;x_m)
    \right\|^2 .
\end{align}
Multiplying \eqref{eq:second_inequality} by $\beta$ and adding it to \eqref{eq:first_inequality}, one cancels the historical gradient terms, yielding
\begin{align}
    &F(\boldsymbol{\theta}_{k+1})-F(\boldsymbol{\theta}_k)
    -\beta
    \left(
    F(\boldsymbol{\theta}_k)-F(\boldsymbol{\theta}_{k-1})
    \right)
    \nonumber\\
    \le&
    -\gamma_k
    \left\langle
    \nabla F(\boldsymbol{\theta}_k),
    \nabla f(\boldsymbol{\theta}_k;x_k)
    \right\rangle
    +
    \frac{L}{2}
    \left\|
    \sum_{m=0}^k
    \beta^{k-m}\gamma_m
    \nabla f(\boldsymbol{\theta}_m;x_m)
    \right\|^2
    \nonumber\\
    &+
    \frac{L\beta}{2}
    \left\|
    \sum_{m=0}^{k-1}
    \beta^{k-1-m}\gamma_m
    \nabla f(\boldsymbol{\theta}_m;x_m)
    \right\|^2 .
\end{align}
Recursively applying the above inequality gives
\begin{align}
\label{eq:main_formula_lemma_nonconvex}
    &F(\boldsymbol{\theta}_{k+1})-F(\boldsymbol{\theta}_k)
    -\beta^k
    \left(
    F(\boldsymbol{\theta}_1)-F(\boldsymbol{\theta}_0)
    \right)
    \nonumber\\
    \le&
    -\sum_{m=1}^k
    \beta^{k-m}\gamma_m
    \left\langle
    \nabla F(\boldsymbol{\theta}_m),
    \nabla f(\boldsymbol{\theta}_m;x_m)
    \right\rangle
    +
    \frac{L}{2}
    \sum_{n=1}^k
    \beta^{k-n}
    \left\|
    \sum_{m=0}^n
    \beta^{n-m}\gamma_m
    \nabla f(\boldsymbol{\theta}_m;x_m)
    \right\|^2
    \nonumber\\
    &+
    \frac{L\beta}{2}
    \sum_{n=1}^k
    \beta^{k-n}
    \left\|
    \sum_{m=0}^{n-1}
    \beta^{n-1-m}\gamma_m
    \nabla f(\boldsymbol{\theta}_m;x_m)
    \right\|^2 .
\end{align}
For the initial step, the descent lemma gives
\begin{align}
\label{eq:first_step_bound_lemma_nonconvex}
    F(\boldsymbol{\theta}_1)-F(\boldsymbol{\theta}_0)
    \le
    -\gamma_0
    \left\langle
    \nabla F(\boldsymbol{\theta}_0),
    \nabla f(\boldsymbol{\theta}_0;x_0)
    \right\rangle
    +
    \frac{L}{2}
    \left\|
    \gamma_0
    \nabla f(\boldsymbol{\theta}_0;x_0)
    \right\|^2 .
\end{align}
Substituting \eqref{eq:first_step_bound_lemma_nonconvex} into \eqref{eq:main_formula_lemma_nonconvex} and collecting the weighted terms proves the lemma.

%% file: Appendix/b.proof_of_theorems.tex
\section{Proofs for standard SHB-PS and SHB-ALS in Section~\ref{sec:main_results}}
\label{appendix:proof_of_theorems}

\input{Main_proofs/proof_of_thm1}

\input{Main_proofs/proof_of_thm2}

\input{Main_proofs/proof_of_thm3}

\input{Main_proofs/proof_of_thm4}

\input{Main_proofs/proof_of_thm5}

%% file: Main_proofs/proof_of_thm1.tex
\subsection{Proof of Theorem~\ref{thm:strongly_convex_general}}
\label{appendix:proof_thm_strongly_convex}

Unrolling the SHB recursion~\eqref{eq:shb} and applying
Lemma~\ref{lemma:decouple_convex}, we have
\begin{align}
    \left\|
        \boldsymbol{\theta}_{k+1}
        -
        \boldsymbol{\theta}^\star
    \right\|^2
    \le&
    \left\|
        \boldsymbol{\theta}_{k}
        -
        \boldsymbol{\theta}^\star
    \right\|^2
    -
    2\sum_{m=0}^k
    \beta^{k-m}\gamma_m
    \left\langle
        \boldsymbol{\theta}_m-\boldsymbol{\theta}^\star,
        \nabla f(\boldsymbol{\theta}_m;x_m)
    \right\rangle
    \nonumber\\
    &+
    2\sum_{n=0}^k
    \beta^{k-n}
    \left\|
        \sum_{m=0}^n
        \beta^{n-m}\gamma_m
        \nabla f(\boldsymbol{\theta}_m;x_m)
    \right\|^2 .
\label{eq:convex_distance_recursion}
\end{align}
By the Cauchy--Schwarz inequality,
\begin{align}
\label{eq:cauchy_schwarz}
\left\|
\sum_{m=0}^n\beta^{n-m}\gamma_m
\nabla f(\boldsymbol{\theta}_m;x_m)
\right\|^2
&\le
\left(
\sum_{m=0}^n
\beta^{\frac{3}{2}(n-m)}
\gamma_m^2
\left\|\nabla f(\boldsymbol{\theta}_m;x_m)\right\|^2
\right)
\left(
\sum_{m=0}^n\beta^{\frac{1}{2}(n-m)}
\right)
\nonumber\\
&\le
\frac{1}{1-\sqrt{\beta}}
\sum_{m=0}^n
\beta^{\frac{3}{2}(n-m)}
\gamma_m^2
\left\|\nabla f(\boldsymbol{\theta}_m;x_m)\right\|^2.
\end{align}
Reordering the summation and applying
Lemma~\ref{lemma:basic_step_size} then gives
\begin{align}
    \sum_{n=0}^k
    \beta^{k-n}
    \left\|
        \sum_{m=0}^n
        \beta^{n-m}\gamma_m
        \nabla f(\boldsymbol{\theta}_m;x_m)
    \right\|^2
    &\le
    \frac{1}{(1-\sqrt{\beta})^2}
    \sum_{m=0}^k
    \beta^{k-m}\gamma_m^2
    \left\|
        \nabla f(\boldsymbol{\theta}_m;x_m)
    \right\|^2
    \nonumber\\
    &\le
    \frac{1}{2c}
    \sum_{m=0}^k
    \beta^{k-m}\gamma_m
    \left(
        f(\boldsymbol{\theta}_m;x_m)-f^\star(x_m)
    \right).
\label{eq:convex_momentum_bound}
\end{align}

Combining~\eqref{eq:convex_distance_recursion}
and~\eqref{eq:convex_momentum_bound}, and using $c\ge1/2$, yields
\begin{align}
\left\|\boldsymbol{\theta}_{k+1}\!-\!\boldsymbol{\theta}^\star\right\|^2
\le&
\left\|\boldsymbol{\theta}_k\!-\!\boldsymbol{\theta}^\star\right\|^2
\!-\!2\sum_{m=0}^k\!\beta^{k-m}\gamma_m
\left(
\left\langle
\boldsymbol{\theta}_m\!-\!\boldsymbol{\theta}^\star,
\nabla f(\boldsymbol{\theta}_m;x_m)
\right\rangle
\!-\!f(\boldsymbol{\theta}_m;x_m)
\!+\!f(\boldsymbol{\theta}^\star;x_m)
\right)
\nonumber\\
&+
2\sum_{m=0}^k\beta^{k-m}\gamma_m
\left(f(\boldsymbol{\theta}^\star;x_m)-f^\star(x_m)\right).
\end{align}
The two summations are nonnegative by convexity and the definition of $f^\star(x_m)$.
Hence, using Lemma~\ref{lemma:step_size_bounds}, taking expectation, and applying the $\mu$-strong convexity of $F$, we obtain
\begin{align}
\mathbb{E}\left\|
\boldsymbol{\theta}_{k+1}-\boldsymbol{\theta}^\star
\right\|^2
\le&
\mathbb{E}\left\|
\boldsymbol{\theta}_k-\boldsymbol{\theta}^\star
\right\|^2
-\mu\gamma_{\min}
\sum_{m=0}^k\beta^{k-m}
\mathbb{E}\left\|
\boldsymbol{\theta}_m-\boldsymbol{\theta}^\star
\right\|^2
+
2\gamma_{\max}\sigma^2
\sum_{m=0}^k\beta^{k-m}.
\label{eq:strongly_convex_expected_recursion}
\end{align}

Let
\begin{align}
e_k
:=
\mathbb{E}\left\|
\boldsymbol{\theta}_k-\boldsymbol{\theta}^\star
\right\|^2
-\frac{2\gamma_{\max}}{\mu\gamma_{\min}}\sigma^2 .
\end{align}
Then~\eqref{eq:strongly_convex_expected_recursion} becomes
\begin{align}
e_{k+1}
\le
e_k-\mu\gamma_{\min}
\sum_{m=0}^k\beta^{k-m}e_m .
\label{eq:strongly_convex_error_recursion}
\end{align}

We first note that once $e_k$ becomes nonpositive, it remains nonpositive thereafter.
For $\beta=0$, this follows immediately from
$e_{k+1}\le(1-\mu\gamma_{\min})e_k$.
For $\beta>0$, let $k_0$ be the first index such that $e_{k_0}\le0$, and define
\begin{align}
S_k:=\sum_{m=k_0}^k\beta^{-m}e_m,
\qquad
S_{k_0-1}:=0.
\end{align}
By~\eqref{eq:strongly_convex_error_recursion},
\begin{align}
\label{eq:Sk_recursion}
\beta S_{k+1}
-\left(\beta+1-\mu\gamma_{\min}\right)S_k
+S_{k-1}
\le0.
\end{align}
Let
\begin{align}
z_\pm
=
\frac{
\beta+1-\mu\gamma_{\min}
\pm
\sqrt{
(\beta+1-\mu\gamma_{\min})^2-4\beta
}
}{2\beta}.
\end{align}
Since
$\mu\gamma_{\min}\le(1-\sqrt{\beta})^2$,
the roots are real and positive, with $z_+>1$.
Thus \eqref{eq:Sk_recursion} gives
\begin{align}
S_{k+1}-z_+S_k
\le
z_-\left(S_k-z_+S_{k-1}\right).
\end{align}
Since $S_{k_0}=\beta^{-k_0}e_{k_0}\le0$, recursively applying the above inequality yields
$S_{k+1}-z_+S_k\le0$ and $S_k\le0$ for all $k\ge k_0$.
Therefore,
\begin{align}
e_{k+1}
=
\beta^{k+1}(S_{k+1}-S_k)
\le0,
\end{align}
which proves the claim.

It remains to consider the case where $e_0,\ldots,e_k>0$.
Retaining only the two most recent terms in~\eqref{eq:strongly_convex_error_recursion} gives
\begin{align}
e_{k+1}
\le
(1-\mu\gamma_{\min})e_k
-\mu\gamma_{\min}\beta e_{k-1}.
\label{eq:strongly_convex_second_order}
\end{align}
Let
\begin{align}
r_\pm
=
\frac{
1-\mu\gamma_{\min}
\pm
\sqrt{
(1-\mu\gamma_{\min})^2
-4\mu\gamma_{\min}\beta
}
}{2}.
\end{align}
The step-size bounds in Lemma~\ref{lemma:step_size_bounds} and $c\ge1/2$ imply
\begin{align}
\mu\gamma_{\min}
\le
\frac{\mu}{2L}(1-\sqrt{\beta})^2
\le
\frac{1}{2}(1-\sqrt{\beta})^2,
\end{align}
which guarantees that the discriminant is positive and
$0\le r_-\le r_+<1$.
Moreover,
$r_++r_-=1-\mu\gamma_{\min}$ and
$r_+r_-=\mu\gamma_{\min}\beta$.
Hence~\eqref{eq:strongly_convex_second_order} can be written as
\begin{align}
e_{k+1}-r_-e_k
\le
r_+\left(e_k-r_-e_{k-1}\right).
\end{align}
Iterating the above inequality gives
\begin{align}
e_k
\le
\frac{e_1-r_-e_0}{r_+-r_-}r_+^k
+
\frac{r_+e_0-e_1}{r_+-r_-}r_-^k
\le
B r_+^k,
\end{align}
for some constant $B\ge0$ determined by the initialization.
Combining this bound with the nonpositive case above, and recalling the definition of $e_k$, yields
\begin{align}
\mathbb{E}\left\|
\boldsymbol{\theta}_k-\boldsymbol{\theta}^\star
\right\|^2
\le
B r^k
+
\frac{2\gamma_{\max}}{\mu\gamma_{\min}}\sigma^2,
\end{align}
where $r=r_+$, which proves the theorem.

%% file: Main_proofs/proof_of_thm2.tex
\subsection{Proof of Theorem~\ref{thm:convex_general}}
\label{appendix:proof_thm_convex}

Following the steps in Appendix~\ref{appendix:proof_thm_strongly_convex}, combining~\eqref{eq:convex_distance_recursion} and~\eqref{eq:convex_momentum_bound}, together with convexity, yields
\begin{align}
    \left\|
        \boldsymbol{\theta}_{k+1}-\boldsymbol{\theta}^\star
    \right\|^2
    \le&
    \left\|
        \boldsymbol{\theta}_{k}-\boldsymbol{\theta}^\star
    \right\|^2
    -
    \left(2-\frac{1}{c}\right)
    \sum_{m=0}^k
    \beta^{k-m}\gamma_m
    \left(
        f(\boldsymbol{\theta}_m;x_m)-f^\star(x_m)
    \right)
    \nonumber\\
    &+
    2\sum_{m=0}^k
    \beta^{k-m}\gamma_m
    \left(
        f(\boldsymbol{\theta}^\star;x_m)-f^\star(x_m)
    \right).
\end{align}
Since $c>1/2$, using the step-size bounds in
Lemma~\ref{lemma:step_size_bounds} gives
\begin{align}
    \left\|
        \boldsymbol{\theta}_{k+1}-\boldsymbol{\theta}^\star
    \right\|^2
    \le&
    \left\|
        \boldsymbol{\theta}_{k}-\boldsymbol{\theta}^\star
    \right\|^2
    -
    \left(2-\frac{1}{c}\right)\gamma_{\min}
    \sum_{m=0}^k
    \beta^{k-m}
    \left(
        f(\boldsymbol{\theta}_m;x_m)
        -
        f(\boldsymbol{\theta}^\star;x_m)
    \right)
    \nonumber\\
    &+
    \left[
        2\gamma_{\max}
        -
        \left(2-\frac{1}{c}\right)\gamma_{\min}
    \right]
    \sum_{m=0}^k
    \beta^{k-m}
    \left(
        f(\boldsymbol{\theta}^\star;x_m)-f^\star(x_m)
    \right).
\label{eq:convex_pathwise_recursion}
\end{align}
Taking expectation and using
Assumption~\ref{assump:finite_optimal_difference}, we obtain
\begin{align}
    \left(2-\frac{1}{c}\right)\gamma_{\min}
    \sum_{m=0}^k
    \beta^{k-m}
    \mathbb{E}
    \left[
        F(\boldsymbol{\theta}_m)-F(\boldsymbol{\theta}^\star)
    \right]
    &\le
    \mathbb{E}
    \left\|
        \boldsymbol{\theta}_{k}-\boldsymbol{\theta}^\star
    \right\|^2
    -
    \mathbb{E}
    \left\|
        \boldsymbol{\theta}_{k+1}-\boldsymbol{\theta}^\star
    \right\|^2
    \nonumber\\
    &\qquad\quad+
    \frac{
        2\gamma_{\max}
        -
        \left(2-\frac{1}{c}\right)\gamma_{\min}
    }{1-\beta}
    \sigma^2 .
\label{eq:convex_expected_recursion}
\end{align}
Retaining the two most recent terms in the weighted sum
and summing~\eqref{eq:convex_expected_recursion} over the iterations
gives
\begin{align}
    &(1+\beta)
    \sum_{m=0}^{k-1}
    \mathbb{E}
    \left[
        F(\boldsymbol{\theta}_m)-F(\boldsymbol{\theta}^\star)
    \right]
\le
    \frac{
        \left\|
            \boldsymbol{\theta}_0-\boldsymbol{\theta}^\star
        \right\|^2
    }{
        \left(2-\frac{1}{c}\right)\gamma_{\min}
    }
    +
    (k+1)
    \left(
        \frac{
            2\gamma_{\max}
        }{
            \left(2-\frac{1}{c}\right)\gamma_{\min}
        }
        -1
    \right)
    \frac{\sigma^2}{1-\beta}.
\label{eq:convex_cumulative_bound}
\end{align}
Finally, by convexity of $F$ and Jensen's inequality,
\begin{align}
    F(\bar{\boldsymbol{\theta}}_k)
    -
    F(\boldsymbol{\theta}^\star)
    \le
    \frac{1}{k}
    \sum_{m=0}^{k-1}
    \left(
        F(\boldsymbol{\theta}_m)-F(\boldsymbol{\theta}^\star)
    \right).
\end{align}
Combining this with~\eqref{eq:convex_cumulative_bound} yields
\begin{align}
    \mathbb{E}
    \left[
        F(\bar{\boldsymbol{\theta}}_k)
        -
        F(\boldsymbol{\theta}^\star)
    \right]
    \le
    \frac{A}{k}
    +
    M\frac{k+1}{k}\sigma^2,
\end{align}
with $A$ and $M$ defined in Theorem~\ref{thm:convex_general},
which completes the proof.

%% file: Main_proofs/proof_of_thm3.tex
\subsection{Proof of Theorem~\ref{thm:nonconvex_general}}
\label{appendix:proof_thm_non_convex}

By Lemma~\ref{lemma:decouple_nonconvex}, we have
\begin{align}
F(\boldsymbol{\theta}_{k+1})-F(\boldsymbol{\theta}_k)
\le&
-\sum_{m=0}^k\beta^{k-m}\gamma_m
\left\langle
\nabla F(\boldsymbol{\theta}_m),
\nabla f(\boldsymbol{\theta}_m;x_m)
\right\rangle \nonumber\\
&+
L\sum_{n=0}^k\beta^{k-n}
\left\|
\sum_{m=0}^n\beta^{n-m}\gamma_m
\nabla f(\boldsymbol{\theta}_m;x_m)
\right\|^2.
\label{eq:nonconvex_descent}
\end{align}
For the first term, using
$\gamma_{\min}\le\gamma_m\le\gamma_{\max}$ gives
\begin{align}
-\gamma_m
\left\langle
\nabla F(\boldsymbol{\theta}_m),
\nabla f(\boldsymbol{\theta}_m;x_m)
\right\rangle
\le&
\frac{\gamma_{\max}-\gamma_{\min}}{2}
\left(
\left\|\nabla F(\boldsymbol{\theta}_m)\right\|^2
+
\left\|\nabla f(\boldsymbol{\theta}_m;x_m)\right\|^2
\right)
\nonumber\\
&-
\gamma_{\max}
\left\langle
\nabla F(\boldsymbol{\theta}_m),
\nabla f(\boldsymbol{\theta}_m;x_m)
\right\rangle .
\label{eq:nonconvex_inner_bound}
\end{align}
For the second term, using~\eqref{eq:cauchy_schwarz} and reordering the summation gives
\begin{align}
L\sum_{n=0}^k\beta^{k-n}
\left\|
\sum_{m=0}^n\beta^{n-m}\gamma_m
\nabla f(\boldsymbol{\theta}_m;x_m)
\right\|^2
&\!\le
\frac{L}{1\!-\!\sqrt{\!\beta}}
\!\sum_{m=0}^k
\beta^{k-m}\gamma_m^2
\left\|\nabla f(\boldsymbol{\theta}_m;x_m)\right\|^2
\!\sum_{n=m}^k\!\beta^{\frac{1}{2}(n-m)}
\nonumber\\
&\le
\frac{L\gamma_{\max}^2}{\left(1\!-\!\sqrt{\!\beta}\right)^2}
\sum_{m=0}^k\beta^{k-m}
\left\|\nabla f(\boldsymbol{\theta}_m;x_m)\right\|^2.
\label{eq:nonconvex_momentum_bound}
\end{align}
Combining~\eqref{eq:nonconvex_descent}, \eqref{eq:nonconvex_inner_bound} and \eqref{eq:nonconvex_momentum_bound},
taking expectation, and applying Assumption~\ref{assump:wgc}, we obtain
\begin{align}
\mathbb{E}
\left[
F(\boldsymbol{\theta}_{k+1})-F(\boldsymbol{\theta}_k)
\right]
\le
-s_1\sum_{m=0}^k\beta^{k-m}
\mathbb{E}
\left\|\nabla F(\boldsymbol{\theta}_m)\right\|^2
+s_2\delta,
\label{eq:nonconvex_expected_descent}
\end{align}
where $s_1$ and $s_2$ are defined in
Theorem~\ref{thm:nonconvex_general}.
The stated hyperparameter conditions ensure $s_1>0$, and we verify this at the end of the proof.
Summing~\eqref{eq:nonconvex_expected_descent} over $k$ and using
$F(\boldsymbol{\theta})\ge F(\boldsymbol{\theta}^\star)$ gives
\begin{align}
s_1
\sum_{n=0}^k\sum_{m=0}^n
\beta^{n-m}
\mathbb{E}
\left\|\nabla F(\boldsymbol{\theta}_m)\right\|^2
\le
F(\boldsymbol{\theta}_0)-F(\boldsymbol{\theta}^\star)
+(k+1)s_2\delta.
\end{align}
Retaining the terms with weights $1$ and $\beta$ yields
\begin{align}
\label{eq:nonconvex_cumulative_bound}
(1+\beta)s_1
\sum_{m=0}^{k-1}
\mathbb{E}
\left\|\nabla F(\boldsymbol{\theta}_m)\right\|^2
\le
F(\boldsymbol{\theta}_0)-F(\boldsymbol{\theta}^\star)
+(k+1)s_2\delta.
\end{align}
Therefore,
\begin{align}
\min_{0\le m\le k-1}
\mathbb{E}
\left\|\nabla F(\boldsymbol{\theta}_m)\right\|^2
\le
\frac{F(\boldsymbol{\theta}_0)-F(\boldsymbol{\theta}^\star)}
{k(1+\beta)s_1}
+
\frac{k+1}{k}
\frac{s_2\delta}{(1+\beta)s_1},
\end{align}
which gives the desired result.

It remains to verify that $s_1>0$ under the stated hyperparameter conditions.
For SHB-ALS, 
we consider the two possible cases. If
$\gamma_{\max}\le (1-\sqrt{\beta})^2\omega(1-c)/L$, then
$\gamma_{\min}=\gamma_{\max}$ and
\begin{align}
    s_1
    =
    \gamma_{\max}
    -
    \rho\frac{L\gamma_{\max}^2}{(1-\sqrt{\beta})^2}
    =
    \gamma_{\max}
    \left(
    1-\frac{\rho L\gamma_{\max}}{(1-\sqrt{\beta})^2}
    \right).
\end{align}
Thus $c>1-\frac{1}{\omega\rho}$ implies $s_1>0$.
Otherwise,
$\gamma_{\min}=(1-\sqrt{\beta})^2\omega(1-c)/L$. Substituting it into the definition of $s_1$ gives
\begin{align}
    2s_1
    =
    -\frac{2\rho L}{(1-\sqrt{\beta})^2}\gamma_{\max}^2
    +(1-\rho)\gamma_{\max}
    +(1+\rho)(1-\sqrt{\beta})^2\frac{\omega(1-c)}{L}.
\end{align}
The right-hand side is positive whenever $\gamma_{\max}$ is smaller than the positive root of the corresponding quadratic equation, namely,
\begin{align}
    \gamma_{\max}
    <
    \frac{(1-\sqrt{\beta})^2}{4\rho L}
    \left(
    1-\rho+
    \sqrt{
    (1-\rho)^2
    +8\rho(1+\rho)\omega(1-c)
    }
    \right),
\end{align}
which is precisely the assumed upper bound for SHB-ALS.

For SHB-PS, similarly,
if $\gamma_{\max}\le (1-\sqrt{\beta})^2/(4cL)$, then again
$\gamma_{\min}=\gamma_{\max}$ and
\begin{align}
    s_1
    =
    \gamma_{\max}
    \left(
    1-\frac{\rho L\gamma_{\max}}{(1-\sqrt{\beta})^2}
    \right).
\end{align}
Moreover, $c\ge\rho/4$ implies
\begin{align}
    \frac{(1-\sqrt{\beta})^2}{4cL}
    \le
    \frac{(1-\sqrt{\beta})^2}{\rho L}.
\end{align}
Together with the strict upper bound on $\gamma_{\max}$ in the theorem, this gives $s_1>0$.
Otherwise,
$\gamma_{\min}=(1-\sqrt{\beta})^2/(4cL)$, and hence
\begin{align}
    2s_1
    =
    -\frac{2\rho L}{(1-\sqrt{\beta})^2}\gamma_{\max}^2
    +(1-\rho)\gamma_{\max}
    +(1+\rho)\frac{(1-\sqrt{\beta})^2}{4cL}.
\end{align}
Therefore, $s_1>0$ whenever
\begin{align}
    \gamma_{\max}
    <
    \frac{(1-\sqrt{\beta})^2}{4\rho L}
    \left(
    1-\rho+
    \sqrt{
    (1-\rho)^2
    +\frac{2\rho(1+\rho)}{c}
    }
    \right),
\end{align}
which is the assumed upper bound for SHB-PS.

%% file: Main_proofs/proof_of_thm4.tex
\subsection{Proof of Theorem~\ref{thm:convex_as}}
\label{appendix:proof_thm_as_convex}

Under interpolation, $\sigma^2=0$. Then following~\eqref{eq:convex_cumulative_bound} in the proof of Theorem~\ref{thm:convex_general}, we have
\begin{align}
    (1+\beta)\sum_{m=0}^{k-1}
    \mathbb{E}\left[
    F(\boldsymbol{\theta}_m)-F(\boldsymbol{\theta}^\star)
    \right]
    \le
    \frac{
    \left\|\boldsymbol{\theta}_0-\boldsymbol{\theta}^\star\right\|^2
    }{
    \left(2-\frac{1}{c}\right)\gamma_{\min}
    }.
\end{align}
Since $F(\boldsymbol{\theta}_m)-F(\boldsymbol{\theta}^\star)\ge0$,
letting $k\to\infty$ and applying the monotone convergence theorem gives
\begin{align}
    \mathbb{E}\left[
    \sum_{m=0}^{\infty}
    \left(
    F(\boldsymbol{\theta}_m)-F(\boldsymbol{\theta}^\star)
    \right)
    \right]
    <\infty.
\end{align}
Hence,
\begin{align}
    \sum_{m=0}^{\infty}
    \left(
    F(\boldsymbol{\theta}_m)-F(\boldsymbol{\theta}^\star)
    \right)
    <\infty
    \quad \mathrm{a.s.},
\end{align}
which implies
$F(\boldsymbol{\theta}_k)\to F(\boldsymbol{\theta}^\star)$ almost surely.
Moreover, by convexity,
\begin{align}
    F(\bar{\boldsymbol{\theta}}_k)
    -F(\boldsymbol{\theta}^\star)
    \le
    \frac{1}{k}
    \sum_{m=0}^{k-1}
    \left(
    F(\boldsymbol{\theta}_m)-F(\boldsymbol{\theta}^\star)
    \right)
    =
    \mathcal{O}(1/k)
    \quad \mathrm{a.s.},
\end{align}
which completes the proof.

%% file: Main_proofs/proof_of_thm5.tex
\subsection{Proof of Theorem~\ref{thm:nonconvex_as}}
\label{appendix:proof_thm_as_non_convex}

When $\delta=0$, using~\eqref{eq:nonconvex_cumulative_bound} in the proof of Theorem~\ref{thm:nonconvex_general} yields
\begin{align}
    \sum_{m=0}^{k-1}
    \mathbb{E}
    \left\|
    \nabla F(\boldsymbol{\theta}_m)
    \right\|^2
    \le
    \frac{
    F(\boldsymbol{\theta}_0)-F(\boldsymbol{\theta}^\star)
    }{
    (1+\beta)s_1
    }.
\end{align}
Letting $k\to\infty$ and applying the monotone convergence theorem yields
\begin{align}
    \mathbb{E}\left[
    \sum_{m=0}^{\infty}
    \left\|
    \nabla F(\boldsymbol{\theta}_m)
    \right\|^2
    \right]
    <\infty.
\end{align}
Therefore,
\begin{align}
    \sum_{m=0}^{\infty}
    \left\|
    \nabla F(\boldsymbol{\theta}_m)
    \right\|^2
    <\infty
    \quad \mathrm{a.s.},
\end{align}
and consequently
$\|\nabla F(\boldsymbol{\theta}_k)\|\to0$ almost surely.
Furthermore,
\begin{align}
    \min_{0\le m\le k-1}
    \left\|
    \nabla F(\boldsymbol{\theta}_m)
    \right\|^2
    \le
    \frac{1}{k}
    \sum_{m=0}^{k-1}
    \left\|
    \nabla F(\boldsymbol{\theta}_m)
    \right\|^2
    =
    \mathcal{O}(1/k)
    \quad \mathrm{a.s.},
\end{align}
which proves the result.

%% file: Appendix/c.proof_of_theorems_diminishing.tex
\section{Proofs for diminishing step sizes in Section~\ref{sec:diminishing}}
\label{appendix:diminishing}
% \textcolor{red}{TODO: All proofs in this Section should be checked and rewritten}

We first collect the auxiliary lemmas used in the diminishing-step-size analysis in Appendix~\ref{appendix:dec_lemmas}.
Then we prove Theorems~\ref{thm:convex_as_dec} and~\ref{thm:nonconvex_as_dec} in Appendices~\ref{appendix:proof_thm_as_convex_dec} and~\ref{appendix:proof_thm_as_nonconvex_dec}, respectively.
Finally, Appendices~\ref{appendix:convex_little_o} and~\ref{appendix:nonconvex_little_o} establish the corresponding little-$o$ refinements.

Throughout this section, let $\mathcal{F}_k=\sigma(\boldsymbol{\theta}_0,x_0,\ldots,x_{k-1})$ denote the natural filtration, i.e., the information before sampling $x_k$.
The schedule hyperparameters are deterministic, with $\eta_k,\gamma_{\max,k}>0$.
All almost sure bounds below allow finite constants that depend on the sample path.
% For a fixed initialization, smoothness and the gradient moment bounds in the respective settings ensure integrability at every finite iteration.

\subsection{Auxiliary lemmas}
\label{appendix:dec_lemmas}

The following is the diminishing-step-size counterpart of Lemma~\ref{lemma:basic_step_size}.
\begin{lemma}
\label{lemma:dec_basic}
For either $\text{SHB\text{-}PS}_{\text{dec}}$ with $c>0$ or
$\text{SHB\text{-}ALS}_{\text{dec}}$ with $c\in(0,1)$, we have
\begin{align}
    \gamma_k\|\nabla f(\boldsymbol{\theta}_k;x_k)\|^2
    \le \frac{\eta_k}{c}
    \bigl(f(\boldsymbol{\theta}_k;x_k)-f^\star(x_k)\bigr).
\end{align}
\end{lemma}

We also use the almost-supermartingale convergence theorem of~\citet{robbins1971convergence}.
\begin{lemma}[Robbins--Siegmund]
\label{lemma:dec_robbins_siegmund}
Let $\{V_k\}$, $\{U_k\}$, and $\{Z_k\}$ be nonnegative, integrable processes adapted to $\{\mathcal{F}_k\}$.
Suppose that $\{\alpha_k\}$ is a deterministic nonnegative sequence with
$\sum_k\alpha_k<\infty$, $\sum_k Z_k<\infty$ almost surely, and
\begin{align}
    \mathbb{E}[V_{k+1}\mid\mathcal{F}_k]+U_k
    \le (1+\alpha_k)V_k+Z_k
    \quad \text{a.s.}
\end{align}
Then $V_k$ converges and $\sum_k U_k<\infty$ almost surely.
\end{lemma}

% The next lemma converts weighted summability into a best-iterate rate.
% \begin{lemma}
% \label{lemma:dec_weighted_minimum}
% Let $\{A_k\}$ and $\{B_k\}$ be deterministic nonnegative sequences such that
% $\sum_k A_k=\infty$, $\sum_k B_k<\infty$, and $A_k-B_k>0$ for all $k\ge k_0$ with sufficient large $k_0$.
% Let $\{C_k\}$ be finite nonnegative random variables satisfying
% \begin{align}
%     \sum_{k=k_0}^{\infty}(A_k-B_k)C_k<\infty
%     \quad \text{a.s.}
% \end{align}
% Then
% \begin{align}
%     \min_{1\le m\le k}C_m
%     =\mathcal{O}\left(\frac{1}{\sum_{m=1}^k A_m}\right)
%     \quad \text{a.s.}
% \end{align}
% \end{lemma}
% \paragraph{Proof.}
% For $k\ge k_0$,
% \begin{align}
%     \left(\min_{1\le j\le k}C_j\right)
%     \sum_{m=k_0}^k(A_m-B_m)
%     \le \sum_{m=k_0}^{\infty}(A_m-B_m)C_m<\infty
%     \quad\mathrm{a.s.}
% \end{align}
% Since $\sum_{m=1}^k A_m\to\infty$ and $\sum_m B_m<\infty$,
% $\sum_{m=k_0}^k(A_m-B_m)\sim\sum_{m=1}^k A_m$.
% The claim follows.

\input{Main_proofs/proof_of_thm6}

\input{Main_proofs/proof_of_thm7}

\input{Main_proofs/proof_of_thm8}

\input{Main_proofs/proof_of_thm9}

%% file: Main_proofs/proof_of_thm6.tex
\subsection{Proof of Theorem~\ref{thm:convex_as_dec}}
\label{appendix:proof_thm_as_convex_dec}

By definition, 
\begin{align}
\label{eq:diminishing_}
&\|\boldsymbol{\theta}_{k+1}\!-\!\boldsymbol{\theta}^\star\|^2\nonumber\\
=&\|\boldsymbol{\theta}_k\!-\!\boldsymbol{\theta}^\star\!-\!\gamma_k\nabla f(\boldsymbol{\theta}_k;x_k)\|^2\!+\!\beta_k^2\|\boldsymbol{\theta}_k\!-\!\boldsymbol{\theta}_{k-1}\|^2\!+\!2\beta_k\langle\boldsymbol{\theta}_{k}\!-\!\boldsymbol{\theta}_{k-1},\boldsymbol{\theta}_k\!-\!\boldsymbol{\theta}^\star\!-\!\gamma_k\nabla f(\boldsymbol{\theta}_k;x_k)\rangle\nonumber\\
\le&(1\!+\!\beta_k)\left(\|\boldsymbol{\theta}_k\!-\!\boldsymbol{\theta}^\star\|^2\!+\!\gamma_k^2\|\nabla\! f(\boldsymbol{\theta}_k;x_k)\|^2\!-\!2\gamma_k\langle\nabla\! f(\boldsymbol{\theta}_k;x_k), \boldsymbol{\theta}_k\!-\!\boldsymbol{\theta}^\star\rangle\!+\!\beta_k\|\boldsymbol{\theta}_k\!-\!\boldsymbol{\theta}_{k-1}\|^2\right).
\end{align}
Combining with convexity and Lemma~\ref{lemma:dec_basic} gives
\begin{align}
    &\frac{\|\boldsymbol{\theta}_{k+1}-\boldsymbol{\theta}^\star\|^2}{1+\beta_k}
    +\gamma_k\left(2-\frac{\eta_k}{c}\right)
       \bigl(f(\boldsymbol{\theta}_k;x_k)-f^\star(x_k)\bigr)
    \nonumber\\
    &\le\|\boldsymbol{\theta}_k-\boldsymbol{\theta}^\star\|^2
       +\beta_k\|\boldsymbol{\theta}_k-\boldsymbol{\theta}_{k-1}\|^2
       +2\gamma_k\bigl(f(\boldsymbol{\theta}^\star;x_k)-f^\star(x_k)\bigr),
    \label{eq:dec_convex_distance}
\end{align}
Similarly, we have
\begin{align}
        &\|\boldsymbol{\theta}_{k+1}-\boldsymbol{\theta}_k\|^2
    +\gamma_k\left(2\beta_k-\frac{\eta_k}{c}\right)
       \bigl(f(\boldsymbol{\theta}_k;x_k)-f^\star(x_k)\bigr)
    \nonumber\\
    &\le\beta_k^2\|\boldsymbol{\theta}_k-\boldsymbol{\theta}_{k-1}\|^2
       +2\beta_k\gamma_k\bigl(f(\boldsymbol{\theta}_{k-1};x_k)-f^\star(x_k)\bigr).
    \label{eq:dec_convex_velocity}
\end{align}
% Here, the first inequality uses
% $\|\boldsymbol{u}+\beta_k\boldsymbol{w}\|^2
% \le(1+\beta_k)(\|\boldsymbol{u}\|^2+\beta_k\|\boldsymbol{w}\|^2)$.
% The second uses convexity to bound the inner product between
% $\nabla f(\boldsymbol{\theta}_k;x_k)$ and $\boldsymbol{\theta}_k-\boldsymbol{\theta}_{k-1}$.

For all sufficiently large $k$, condition~\eqref{eq:condition_c_as_convex} implies
\begin{align}
    \eta_k\gamma_{\min,k}\left(1+\beta_k-\frac{\eta_k}{c}\right)
    \ge\frac{\eta_k\gamma_{\max,k}\beta_k}{1+\beta_k}.
    \label{eq:dec_convex_coefficient}
\end{align}
Adding~\eqref{eq:dec_convex_distance} and~\eqref{eq:dec_convex_velocity}, applying Lemma~\ref{lemma:dec_stepsize_bounds}, and taking conditional expectation yields
\begin{align}
    &\frac{\mathbb{E}[\|\boldsymbol{\theta}_{k+1}-\boldsymbol{\theta}^\star\|^2\mid\mathcal{F}_k]}{1+\beta_k}
     +\mathbb{E}[\|\boldsymbol{\theta}_{k+1}-\boldsymbol{\theta}_k\|^2\mid\mathcal{F}_k]
     +\frac{2\eta_k\gamma_{\max,k}\beta_k}{1+\beta_k}
       \bigl(F(\boldsymbol{\theta}_k)-F(\boldsymbol{\theta}^\star)\bigr)
     \nonumber\\
    &\le\|\boldsymbol{\theta}_k-\boldsymbol{\theta}^\star\|^2
       +\beta_k(1+\beta_k)\|\boldsymbol{\theta}_k-\boldsymbol{\theta}_{k-1}\|^2
     \nonumber+2\eta_k\gamma_{\max,k}\beta_k
       \bigl(F(\boldsymbol{\theta}_{k-1})-F(\boldsymbol{\theta}^\star)\bigr)
     \nonumber\\
    &\quad+\left(2\eta_k(1+\beta_k)(\gamma_{\max,k}-\gamma_{\min,k})
       +\frac{2\eta_k^2\gamma_{\min,k}}{c}\right)\sigma^2.
    \label{eq:dec_convex_combined}
\end{align}
% The coefficient reduction in~\eqref{eq:dec_convex_combined} is made after conditioning, using
% $F(\boldsymbol{\theta}_k)-F(\boldsymbol{\theta}^\star)\ge0$.

Consider the nonnegative Lyapunov sequence
\begin{align}
    V_k={}&\|\boldsymbol{\theta}_k\!-\!\boldsymbol{\theta}^\star\|^2
       +(1+\beta_{k-1})\|\boldsymbol{\theta}_k\!-\!\boldsymbol{\theta}_{k-1}\|^2
    +2\eta_{k-1}\gamma_{\max,k-1}\beta_{k-1}
       \bigl(F(\boldsymbol{\theta}_{k-1})\!-\!F(\boldsymbol{\theta}^\star)\bigr).
\end{align}
The monotonicity of $\eta_k\gamma_{\max,k}$ and $\beta_k$, together with~\eqref{eq:dec_convex_combined}, gives
\begin{align}
    &\mathbb{E}[V_{k+1}\mid\mathcal{F}_k]
      +(1-\beta_k)\|\boldsymbol{\theta}_k-\boldsymbol{\theta}_{k-1}\|^2
      \nonumber\\
    &\le(1+\beta_k)V_k
      +(1+\beta_k)\left(2\eta_k(1+\beta_k)(\gamma_{\max,k}-\gamma_{\min,k})
       +\frac{2\eta_k^2\gamma_{\min,k}}{c}\right)\sigma^2.
    \label{eq:dec_convex_lyapunov}
\end{align}
Indeed, the available coefficient of $\|\boldsymbol{\theta}_k-\boldsymbol{\theta}_{k-1}\|^2$ is at least
$(1+\beta_k)(1-\beta_k^2)\ge1-\beta_k$.
By Lemma~\ref{lemma:dec_stepsize_bounds} and $\gamma_{\max,k}\to0$,
$\gamma_{\max,k}-\gamma_{\min,k}$ vanishes for all sufficiently large $k$.
Moreover, $\sum_k\eta_k^2\gamma_{\min,k}\le\sum_k\eta_k^2\gamma_{\max,k}<\infty$.
Thus the error term in~\eqref{eq:dec_convex_lyapunov} is summable.
Since $\sum_k\beta_k<\infty$ and $1-\beta_k$ is bounded away from zero, Lemma~\ref{lemma:dec_robbins_siegmund} implies
\begin{align}
    \sum_{k=1}^{\infty}\|\boldsymbol{\theta}_k-\boldsymbol{\theta}_{k-1}\|^2<\infty
    \quad\mathrm{a.s.}
    \label{eq:dec_convex_velocity_sum}
\end{align}
Returning to~\eqref{eq:dec_convex_distance}, the step-size bounds and conditional expectation give
\begin{align}
    &\mathbb{E}[\|\boldsymbol{\theta}_{k+1}-\boldsymbol{\theta}^\star\|^2\mid\mathcal{F}_k]
       +\eta_k\gamma_{\min,k}(1+\beta_k)\left(2-\frac{\eta_k}{c}\right)
       \bigl(F(\boldsymbol{\theta}_k)-F(\boldsymbol{\theta}^\star)\bigr)
       \nonumber\\
    &\le(1+\beta_k)\|\boldsymbol{\theta}_k-\boldsymbol{\theta}^\star\|^2
       +\beta_k(1+\beta_k)\|\boldsymbol{\theta}_k-\boldsymbol{\theta}_{k-1}\|^2
       \nonumber\\
    &\quad+(1+\beta_k)\left(2\eta_k(\gamma_{\max,k}-\gamma_{\min,k})
       +\frac{\eta_k^2\gamma_{\min,k}}{c}\right)\sigma^2.
    \label{eq:dec_convex_second_rs}
\end{align}
Condition~\eqref{eq:dec_convex_coefficient} implies $\eta_k/c\le1+\beta_k$, and hence
$(1+\beta_k)(2-\eta_k/c)\ge1-\beta_k^2$ for all sufficiently large $k$.
The error terms in~\eqref{eq:dec_convex_second_rs} are summable by~\eqref{eq:dec_convex_velocity_sum} and the same bounds used above.
A second application of Lemma~\ref{lemma:dec_robbins_siegmund} shows that
$\|\boldsymbol{\theta}_k-\boldsymbol{\theta}^\star\|^2$ converges and
\begin{align}
    \sum_{k=1}^{\infty}\eta_k\gamma_{\min,k}
      \bigl(F(\boldsymbol{\theta}_k)-F(\boldsymbol{\theta}^\star)\bigr)<\infty
    \quad\mathrm{a.s.}
    \label{eq:dec_convex_gap_sum}
\end{align}
Finally, Jensen's inequality yields
\begin{align}
    F(\bar{\boldsymbol{\theta}}_k^{\rm w})\!-\!F(\boldsymbol{\theta}^\star)
    &\le\frac{\sum_{m=1}^k\eta_m\gamma_{\min,m}
       \bigl(F(\boldsymbol{\theta}_m)\!-\!F(\boldsymbol{\theta}^\star)\bigr)}
       {\sum_{m=1}^k\eta_m\gamma_{\min,m}}=\mathcal{O}\left(\frac{1}{\sum_{m=1}^k\eta_m\gamma_{\max,m}}\right)
      \quad\mathrm{a.s.},
\end{align}
where $\sum_{m=1}^k\eta_m\gamma_{\min,m}\sim\sum_{m=1}^k\eta_m\gamma_{\max,m}$ by the step-size bounds, and
\begin{align}
\bar{\boldsymbol{\theta}}_k^{\rm w}
    =
    \frac{
        \sum_{m=1}^k
        \eta_m\gamma_{\min,m}\boldsymbol{\theta}_m
    }{
        \sum_{m=1}^k \eta_m\gamma_{\min,m}
    }.
\end{align}
This proves the theorem.

%% file: Main_proofs/proof_of_thm7.tex
\subsection{Proof of Theorem~\ref{thm:nonconvex_as_dec}}
\label{appendix:proof_thm_as_nonconvex_dec}
By smoothness,
\begin{align}
&F({\boldsymbol{\theta}}_{k+1})-F({\boldsymbol{\theta}}_k)\nonumber\\
\le &\langle\nabla F({\boldsymbol{\theta}}_k),{\boldsymbol{\theta}}_{k+1}-{\boldsymbol{\theta}}_k\rangle+\frac{L}{2}\|{\boldsymbol{\theta}}_{k+1}-{\boldsymbol{\theta}}_k\|^2\nonumber\\
=& -\gamma_k\langle\nabla F({\boldsymbol{\theta}}_k), \nabla f({\boldsymbol{\theta}}_k;\mathbf{x}_k)\rangle+\langle\nabla F({\boldsymbol{\theta}}_k), \beta_k({\boldsymbol{\theta}}_k-{\boldsymbol{\theta}}_{k-1})\rangle+\frac{L}{2}\|{\boldsymbol{\theta}}_{k+1}-{\boldsymbol{\theta}}_k\|^2.
\label{eq:nonlinear_diminishing_basic}
\end{align}
We have
\begin{align}
\langle\nabla F({\boldsymbol{\theta}}_k),{\boldsymbol{\theta}}_k-{\boldsymbol{\theta}}_{k-1}\rangle\le F({\boldsymbol{\theta}}_k)-F({\boldsymbol{\theta}}_{k-1})+\frac{L}{2}\|{\boldsymbol{\theta}}_{k}-{\boldsymbol{\theta}}_{k-1}\|^2.\label{eq:nonlinear_diminishing_bound1}
\end{align}
and
\begin{align}
&-\gamma_k\langle\nabla F({\boldsymbol{\theta}}_k),\nabla f({\boldsymbol{\theta}}_k;\mathbf{x}_k)\rangle\nonumber\\
=&-\frac{\gamma_k}{2}\|\nabla F({\boldsymbol{\theta}}_k)\|^2-\frac{\gamma_k}{2}\|\nabla f({\boldsymbol{\theta}}_k;\mathbf{x}_k)\|^2+\frac{\gamma_k}{2}\|\nabla F({\boldsymbol{\theta}}_k) - \nabla f({\boldsymbol{\theta}}_k;\mathbf{x}_k)\|^2\nonumber\\
\le &\frac{\eta_k\gamma_{\max, k}\!-\!\eta_k\gamma_{\min, k}}{2}\left(\|\nabla\! F({\boldsymbol{\theta}}_k)\|^2\!+\!\|\nabla\! f({\boldsymbol{\theta}}_k;\mathbf{x}_k)\|^2\right)\!-\!\eta_k\gamma_{\max, k}\langle\nabla\! F({\boldsymbol{\theta}}_k), \nabla\! f({\boldsymbol{\theta}}_k;\mathbf{x}_k)\rangle.\label{eq:nonlinear_diminishing_bound2}
\end{align}
Insert \eqref{eq:nonlinear_diminishing_bound1} and \eqref{eq:nonlinear_diminishing_bound2} to \eqref{eq:nonlinear_diminishing_basic} and note that $F({\boldsymbol{\theta}}_{k-1})\ge F({\boldsymbol{\theta}}^*)$,
\begin{align}
&F({\boldsymbol{\theta}}_{k+1})-F({\boldsymbol{\theta}}^*)\nonumber\\
\le &(1+\beta_k)(F({\boldsymbol{\theta}}_k)-F({\boldsymbol{\theta}}^*))+\frac{L\beta_k}{2}\|{\boldsymbol{\theta}}_k-{\boldsymbol{\theta}}_{k-1}\|^2+\frac{L}{2}\|{\boldsymbol{\theta}}_{k+1}-{\boldsymbol{\theta}}_k\|^2\nonumber\\
&\!+\!\frac{\eta_k\gamma_{\max, k}\!-\!\eta_k\gamma_{\min, k}}{2}\!\left(\|\nabla\! F({\boldsymbol{\theta}}_k)\|^2\!+\!\|\nabla\! f({\boldsymbol{\theta}}_k;\mathbf{x}_k)\|^2\right)\!-\!\eta_k\gamma_{\max, k}\langle\nabla\! F({\boldsymbol{\theta}}_k),\!\nabla\! f({\boldsymbol{\theta}}_{k};\mathbf{x}_k)\rangle.
\end{align}
Then taking expectation condition on $\mathcal{F}_k$ and applying Assumption~\ref{assump:wgc} yields
\begin{align}
    \mathbb{E}[F(\boldsymbol{\theta}_{k+1})-F(\boldsymbol{\theta}^\star)\mid\mathcal{F}_k]\le&(1+\beta_k)\bigl(F(\boldsymbol{\theta}_k)-F(\boldsymbol{\theta}^\star)\bigr)
       +\frac{L\beta_k}{2}\|\boldsymbol{\theta}_k-\boldsymbol{\theta}_{k-1}\|^2
    \nonumber\\
    &+\frac{L}{2}\mathbb{E}[\|\boldsymbol{\theta}_{k+1}-\boldsymbol{\theta}_k\|^2\mid\mathcal{F}_k]
       +\frac{\eta_k(\gamma_{\max,k}-\gamma_{\min,k})}{2}\delta
    \nonumber\\
    &-\left(\eta_k\gamma_{\max,k}
       -\frac{(1+\rho)\eta_k(\gamma_{\max,k}-\gamma_{\min,k})}{2}\right)
       \|\nabla F(\boldsymbol{\theta}_k)\|^2.
    \label{eq:dec_nonconvex_descent}
\end{align}
We also note that
\begin{align}
\|{\boldsymbol{\theta}}_{k+1}-{\boldsymbol{\theta}}_k\|^2
\le &(\gamma_k^2+\gamma_k\beta_k)\|\nabla f({\boldsymbol{\theta}}_{k};\mathbf{x}_k)\|^2+(\beta_k^2+\gamma_k\beta_k)\|{\boldsymbol{\theta}}_{k}-{\boldsymbol{\theta}}_{k-1}\|^2
\end{align}
Thus, again by Assumption~\ref{assump:wgc},
\begin{align}
    \mathbb{E}[\|{\boldsymbol{\theta}}_{k+1}-{\boldsymbol{\theta}}_k\|^2\mid&\mathcal{F}_k]\le (\eta_k^2\gamma_{\max,k}^2+\eta_k\gamma_{\max, k}\beta_k)\rho\|\nabla F({\boldsymbol{\theta}}_k)\|^2\nonumber\\
    &+(\beta_k^2+\beta_k\eta_k\gamma_{\max, k})\|{\boldsymbol{\theta}}_k-{\boldsymbol{\theta}}_{k-1}\|^2+(\eta_k^2\gamma_{\max,k}^2+\eta_k\gamma_{\max, k}\beta_k)\delta.
        \label{eq:dec_nonconvex_velocity}
\end{align}
Denote
\begin{align}
     &C_k=\frac{\frac{L\beta_k}{2}(1+\beta_k+\eta_k\gamma_{\max, k})}{1+\beta_k(1-\beta_k-\eta_k\gamma_{\max, k})},\nonumber\\
&D_k=\eta_k\gamma_{\max, k}-(1+\rho)\eta_k\frac{\gamma_{\max, k}-\gamma_{\min, k}}{2}-\left(\frac{L}{2}+C_k\right)(\eta_k\gamma_{\max, k}+\beta_k)\eta_k\gamma_{\max, k}\rho,\nonumber\\
&E_k=\frac{\eta_k\gamma_{\max, k}-\eta_k\gamma_{\min, k}}{2}+\left(\frac{L}{2}+C_k\right)(\eta_k\gamma_{\max, k}+\beta_k)\eta_k\gamma_{\max, k}.\nonumber
\end{align}
One may verify that $C_k$ is decreasing under Assumption~\ref{assump:as_decay} and
\begin{align}
    &\mathbb{E}[F({\boldsymbol{\theta}}_{k+1})-F({\boldsymbol{\theta}}^*)+C_k\|{\boldsymbol{\theta}}_{k+1}-{\boldsymbol{\theta}}_{k}\|^2\mid \mathcal{F}_k]+D_k\|\nabla F({\boldsymbol{\theta}}_k)\|^2\nonumber\\
    &\le (1+\beta_k)\left(F({\boldsymbol{\theta}}_{k})-F({\boldsymbol{\theta}}^*)+C_{k-1}\|{\boldsymbol{\theta}}_{k}-{\boldsymbol{\theta}}_{k-1}\|^2\right)+E_k\delta.
\end{align}
The error term $E_k\delta$ is summable because
\begin{align}
    &\sum_k\eta_k^2\gamma_{\max,k}^2
    \le\left(\sup_k\gamma_{\max,k}\right)
       \sum_k\eta_k^2\gamma_{\max,k}<\infty,\nonumber\\
    &\sum_k\eta_k\gamma_{\max,k}\beta_k
    \le\eta_1\gamma_{\max,1}\sum_k\beta_k<\infty,\nonumber
\end{align}
and $\gamma_{\max,k}-\gamma_{\min,k}$ vanishes eventually by Assumption~\ref{assump:as_decay} and Lemma~\ref{lemma:dec_stepsize_bounds}.
And similarly, $D_k\sim\eta_k\gamma_{\max, k}$ and $C_k,D_k>0$ for a sufficiently large $k$.
Hence, applying Lemma~\ref{lemma:dec_robbins_siegmund} gives
\begin{align}
    \sum_{k=1}^{\infty}\eta_k\gamma_{\max,k}
       \|\nabla F(\boldsymbol{\theta}_k)\|^2<\infty
    \quad\mathrm{a.s.}
    \label{eq:dec_nonconvex_gap_sum}
\end{align}
which completes the proof.

%% file: Main_proofs/proof_of_thm8.tex
\subsection{Little-\texorpdfstring{$o$}{o} refinement for convex objectives}
\label{appendix:convex_little_o}

\begin{theorem}
\label{thm:convex_as_dec_little_o}
Under the same conditions as Theorem~\ref{thm:convex_as_dec}, we have
\begin{align}
    F(\bar{\boldsymbol{\theta}}_k^{\rm w})-F(\boldsymbol{\theta}^\star)
    =o\left(\frac{1}{\sum_{m=1}^k\eta_m\gamma_{\min,m}}\right)
    \quad\mathrm{a.s.}
\end{align}
\end{theorem}

\paragraph{Proof.}

The proof of Theorem~\ref{thm:convex_as_dec} has shown that $\|\boldsymbol{\theta}_k-\boldsymbol{\theta}^*\|^2$ converges almost surely and $F(\bar{\boldsymbol{\theta}})\rightarrow F(\boldsymbol{\theta}^\star)$ almost surely, which leads to almost sure boundedness of $\{\boldsymbol{\theta}_k\}$. Define the tail average
\begin{align}
    \boldsymbol{y}_{n,k}
    =\frac{\sum_{m=n+1}^k\eta_m\gamma_{\min,m}\boldsymbol{\theta}_m}
          {\sum_{m=n+1}^k\eta_m\gamma_{\min,m}},\qquad k>n.
\end{align}
Then
\begin{align}
\label{eq:tail_separation}
    \bar{\boldsymbol{\theta}}_k^{\rm w}
    =\frac{\sum_{m=1}^n\eta_m\gamma_{\min,m}}
          {\sum_{m=1}^k\eta_m\gamma_{\min,m}}\bar{\boldsymbol{\theta}}_n^{\rm w}
     +\frac{\sum_{m=n+1}^k\eta_m\gamma_{\min,m}}
           {\sum_{m=1}^k\eta_m\gamma_{\min,m}}\boldsymbol{y}_{n,k}.
\end{align}
Boundedness provides a finite constant $M$ such that
$\|\bar{\boldsymbol{\theta}}_n^{\rm w}-\boldsymbol{y}_{n,k}\|\le M$ for every $k>n$.
Then by smoothness and \eqref{eq:tail_separation},
\begin{align}
    \left(\sum_{m=1}^k\eta_m\gamma_{\min,m}\right)&
       \bigl(F(\bar{\boldsymbol{\theta}}_k^{\rm w})-F(\boldsymbol{\theta}^\star)\bigr)
       \le\left(\sum_{m=1}^k\eta_m\gamma_{\min,m}\right)
       \bigl(F(\boldsymbol{y}_{n,k})-F(\boldsymbol{\theta}^\star)\bigr)
       \nonumber\\
    &+M\left(\sum_{m=1}^n\eta_m\gamma_{\min,m}\right)
       \|\nabla F(\boldsymbol{y}_{n,k})\|
       +\frac{LM^2\left(\sum_{m=1}^n\eta_m\gamma_{\min,m}\right)^2}
       {2\sum_{m=1}^k\eta_m\gamma_{\min,m}}.
\end{align}
Smoothness also yields
\begin{align}
    \|\nabla F(\boldsymbol{y}_{n,k})\|^2\le 2L\left(F(\boldsymbol{y}_{n, k})-F(\boldsymbol{\theta}^\star)\right).
\end{align}
Thus $\lim_{k\rightarrow \infty}\|\nabla F(\boldsymbol{y}_{n, k})\|=0$.
By convexity,
\begin{align}
    F(\boldsymbol{y}_{n,k})-F(\boldsymbol{\theta}^\star)
    \le\frac{\sum_{m=n+1}^{\infty}\eta_m\gamma_{\min,m}
       \bigl(F(\boldsymbol{\theta}_m)-F(\boldsymbol{\theta}^\star)\bigr)}
       {\sum_{m=n+1}^k\eta_m\gamma_{\min,m}}.
    \label{eq:dec_convex_tail_average}
\end{align}
Consequently,
\begin{align}
    \limsup_{k\to\infty}
       \left(\sum_{m=1}^k\eta_m\gamma_{\min,m}\right)
       \bigl(F(\bar{\boldsymbol{\theta}}_k^{\rm w})-F(\boldsymbol{\theta}^\star)\bigr)
       \le\sum_{m=n+1}^{\infty}\eta_m\gamma_{\min,m}
       \bigl(F(\boldsymbol{\theta}_m)-F(\boldsymbol{\theta}^\star)\bigr).
\end{align}
Letting $n\to\infty$ and by Theorem~\ref{thm:convex_as_dec}, the right-hand side goes to zero.
Finally, since $\gamma_{\min,m}=\gamma_{\max,m}$ for all sufficiently large $m$ and $\sum_{m=1}^k\eta_m\gamma_{\min,m}\to\infty$, Theorem~\ref{thm:convex_as_dec_little_o} follows.

%% file: Main_proofs/proof_of_thm9.tex
\subsection{Little-\texorpdfstring{$o$}{o} refinement for non-convex objectives}
\label{appendix:nonconvex_little_o}

\begin{theorem}
\label{thm:nonconvex_as_dec_little_o}
Under the same conditions as Theorem~\ref{thm:nonconvex_as_dec}, we have
\begin{align}
    \min_{1\le m\le k}\|\nabla F(\boldsymbol{\theta}_m)\|^2
    =o\left(\frac{1}{\sum_{m=1}^k\eta_m\gamma_{\max,m}}\right)
    \quad\mathrm{a.s.}
\end{align}
\end{theorem}

\paragraph{Proof.}
For any fixed $n$ and $k\ge n$,
\begin{align}
    \left(\min_{1\le j\le k}\|\nabla F(\boldsymbol{\theta}_j)\|^2\right)
       \sum_{m=1}^k\eta_m\gamma_{\max,m}
       &\le\sum_{m=n}^{\infty}\eta_m\gamma_{\max,m}
       \|\nabla F(\boldsymbol{\theta}_m)\|^2\nonumber\\
       &+\left(\min_{1\le j\le k}\|\nabla F(\boldsymbol{\theta}_j)\|^2\right)\sum_{m=1}^n\eta_m\gamma_{\max,m}.
\end{align}
By Theorem~\ref{thm:nonconvex_as_dec}, we obtain
\begin{align}
    &\limsup_{k\to\infty}
       \left(\sum_{m=1}^k\eta_m\gamma_{\max,m}\right)
       \min_{1\le j\le k}\|\nabla F(\boldsymbol{\theta}_j)\|^2
       \le\sum_{m=n}^{\infty}\eta_m\gamma_{\max,m}
       \|\nabla F(\boldsymbol{\theta}_m)\|^2
       \quad\mathrm{a.s.}
\end{align}
The right-hand side tends to zero as $n\to\infty$ again by Theorem~\ref{thm:nonconvex_as_dec}. Thus Theorem~\ref{thm:nonconvex_as_dec_little_o} follows.

%% file: Appendix/d.more_related_works.tex
\section{More related works}
\label{appendix:more_related_works}

\subsection{Stochastic Polyak step sizes with momentum}

Within the literature on Polyak-type step sizes for stochastic optimization, several works have explored their combination with momentum motivated by the practical benefits of momentum acceleration.
Early approaches such as L4Mom~\citep{rolinek2018l4} and ALI-G~\citep{berrada2020training} incorporated momentum empirically, while their convergence analyses do not cover the corresponding momentum variants.

More recent works have started to investigate the convergence theory of Polyak-type step sizes with momentum.
Among the earliest such analyses, ALR-SMAG~\citep{ALR-SMAG} considers stochastic moving averaged gradients (SMAG), whose update takes the form
\begin{align}
    \boldsymbol{\theta}_{k+1}
    =
    \boldsymbol{\theta}_k
    -
    \gamma_k
    \sum_{m=0}^k
    \beta^{k-m}
    \nabla f(\boldsymbol{\theta}_m;x_m).
\end{align}
In SMAG, the current step size $\gamma_k$ is applied to all historical gradients accumulated in the momentum direction.
Consequently, SMAG and SHB are equivalent under a constant step size but generally differ when adaptive step sizes are used, as also noted in~\citep{ALR-SMAG}.
Therefore, the convergence guarantees developed for ALR-SMAG do not directly apply to SHB.
Notably,~\citet{ALR-SMAG} also introduce Polyak step-size variants for SHB and evaluate them empirically, but do not establish corresponding convergence guarantees.
SGDM-APS~\citep{SGDM-APS} further develops a momentum-aware Polyak step size for SMAG and establishes convergence rates for both convex and non-convex objectives.
Our results complement SGDM-APS~\citep{SGDM-APS} by establishing convergence guarantees for Polyak step size in SHB, alongside their analysis for SMAG.
% Their results and ours are thus complementary, providing general convergence analyses for Polyak-type adaptive step sizes with momentum under the SMAG and SHB formulations, respectively.

Among the existing works, MomSPS~\citep{MomSPS} is most closely related to ours, as it directly studies Polyak step sizes for SHB.
Convergence analysis of $\text{MomSPS}_\text{max}$~\citep{MomSPS} focuses on convex objectives and imposes restrictions on the momentum parameter $\beta$.
In comparison, our results for standard SHB-PS allow any $\beta\in[0,1)$ in the convex setting, cover general non-convex objectives, and further establish almost sure convergence under interpolation.

\subsection{Stochastic line search with momentum}

SLS~\citep{sls} already incorporated momentum into stochastic line search as a practical acceleration technique.
In particular, it considered both heavy-ball momentum and Nesterov acceleration, and demonstrated empirically that momentum can substantially accelerate convergence, while its convergence analysis is restricted to the momentum-free setting.
MSL~\citep{fan2023msl} further explored stochastic line search with momentum-based search directions.
It introduced momentum correction and restart mechanisms to address the problem that a momentum direction may not be a descent direction for the current stochastic objective, with its main focus on empirical performance.

More recently, Lapucci and Pucci~\citep{lapucci2025convergence,lapucci2026effectively} have studied convergence guarantees for stochastic line searches with more general search directions.
\citet{lapucci2025convergence} characterize conditions on the search direction that ensure a well-defined stochastic line search and establish fast convergence.
The subsequent work~\citep{lapucci2026effectively} explicitly incorporates momentum-like directions into the line-search framework, using a dynamically selected momentum parameter $\beta_k$ based on conjugate gradient rules, and establishes convergence guarantees under interpolation and the PL condition.
In comparison, for the non-diminishing variants, our guarantees allow any fixed
$\beta\in[0,1)$ and cover strongly convex, convex, and general
non-convex objectives without requiring interpolation or the
PL condition.
This setting is also closer to the standard use of momentum in practice, where $\beta$ is typically kept fixed throughout optimization.
Furthermore, under suitable step-size and momentum decay conditions,
we establish almost sure convergence to the optimum or stationary beyond
interpolation.
% In comparison, our analysis considers the canonical SHB update with a fixed momentum parameter $\beta$ and covers both non-interpolation and general non-convex settings.
% %
% This setting is also closer to the standard use of momentum in practice, where $\beta$ is typically kept fixed throughout optimization.

\subsection{Diminishing variants and exact convergence}

A number of works have studied diminishing or adaptive variants of PS and ALS to recover exact convergence beyond the interpolation regime.
Most of this literature focuses on momentum-free stochastic optimization.
For stochastic line-search methods, \citet{bellavia2026slises} study a subsampled line-search spectral-gradient scheme with diminishing step sizes and establish almost sure convergence guarantees.
For Polyak-type methods, \citet{sps_dec} introduce DecSPS, a diminishing variant of SPS that converges to the exact solution beyond interpolation, with convergence guarantees given in expectation.
\citet{jiang2023adaptive} further propose AdaSPS and AdaSLS, which adapt to both interpolation and non-interpolation regimes, and additionally combine these methods with variance reduction to obtain improved expected convergence rates.
Most closely related to our setting, \citet{sebbouh2021almost} analyze decreasing variants of both ALS and PS for SGD and establish almost sure convergence rates for convex objectives.
%
% Their construction is particularly close to the diminishing step-size rules considered here.

For momentum methods, \citet{MomSPS} extend decreasing Polyak step sizes to SHB through MomDecSPS and MomAdaSPS, and establish expected convergence to the exact minimizer for convex objectives without interpolation.
Our results complement these works in two aspects.
First, we establish almost sure convergence for diminishing PS and ALS.
Second, our analysis covers not only convex objectives, but also general non-convex objectives, for which we prove almost sure convergence to stationarity.
In particular, while the diminishing PS and ALS considered in our analysis are closely related to \citet{sebbouh2021almost} for SGD, we extend their almost-sure perspective to the stochastic heavy ball setting and further provide guarantees for general non-convex objectives.

%% file: Appendix/e.numerical_proof_of_concepts.tex
\section{Additional numerical results}
\label{appendix:numerical}

\begin{figure}[t]
    \centering
    \includegraphics[width=\textwidth]{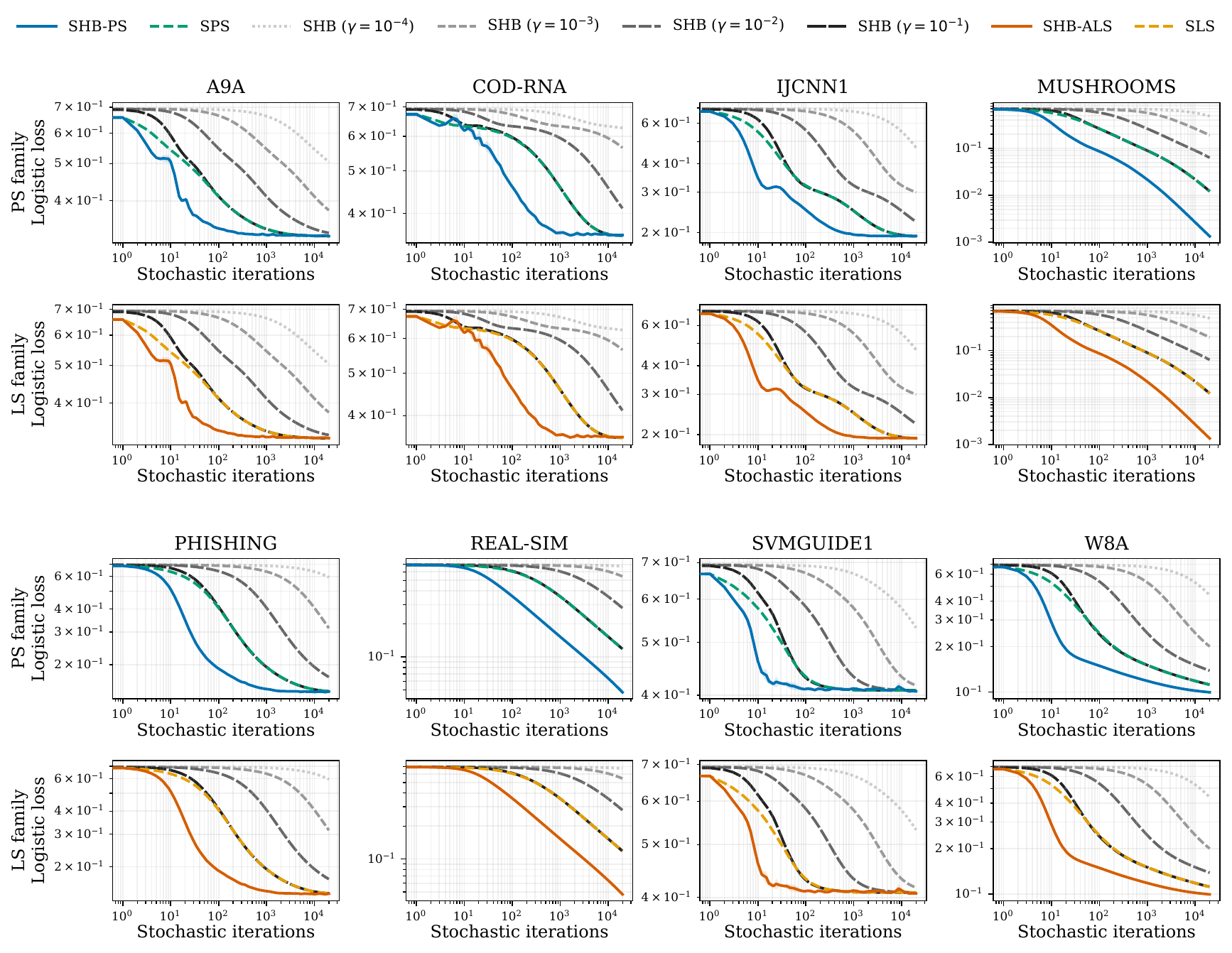}
    \caption{Full logistic training loss on eight LIBSVM datasets. 
    % Within
    % each four-dataset block, the upper and lower rows show the PS and LS
    % families, respectively.
    }
    \label{fig:svm_numerical}
    \vspace{-1em}
\end{figure}

Here we present more results on practical machine learning tasks.
We emphasize that the goal of this work is not to propose new algorithms, but to provide more general convergence guarantees for existing adaptive step-size methods with momentum.
% Accordingly, our numerical experiments are intended to examine whether the predicted convergence behavior can be observed on practical optimization tasks and whether several implications of our theory are supported empirically.
Accordingly, these experiments are intended to complement our theoretical analysis by illustrating the practical behavior of the studied methods on representative machine learning tasks, rather than to establish their empirical advantage.
% All curves in this appendix report the mean and $95\%$ confidence interval over five random seeds.
The curves for logistic-regression experiments on LIBSVM datasets report the mean and 95\% confidence interval over five random seeds, while the deep learning experiments use three seeds.

\paragraph{Practical step-size rescaling.}
In the literature on PS and ALS, numerical evaluations often employ larger effective step sizes than those directly suggested by the theoretical analysis~\citep{sls,SPS,ALR-SMAG,MomSPS,SGDM-APS}.
For Polyak-type methods, this can be conveniently achieved by choosing a smaller value of $c$~\citep{SPS, SGDM-APS}, whereas the same adjustment is not directly applicable to ALS.
We follow this practical convention in our experiments.
To obtain comparable rescaling for both SHB-PS and SHB-ALS, we multiply the adaptive step size by a $\beta$-dependent factor $r_\beta$.

\paragraph{Hyperparameters.}
We use $\beta=0.9$ and set
$c=1$ for SHB-PS and $c=0.1$ for SHB-ALS.
Both methods use the common rescaling factor
$r_\beta=(1-\sqrt{\beta})^{-2}$.
We cap the step size after rescaling:
\begin{align}
\gamma_k^{\mathrm{eff}}
=\min\{r_\beta\gamma_k^{\mathrm{raw}},\gamma_{\max}\},
\qquad \gamma_{\max}=1,
\end{align}
where $\gamma_k^{\mathrm{raw}}$ denotes the adaptive step size
before rescaling.
The same rescaling rule and effective step-size cap are used throughout this appendix.
We adopt the smooth reset strategy~\citep{sls,SPS}
with $q=2^{2/n_b}$ for SHB-PS and $q=2^{1/n_b}$ for SHB-ALS, where $n_b$ denotes the number of minibatches
per epoch.
The initial raw step-size proposal is $\gamma_{\max}/r_\beta$,
and subsequent proposals are
$\min\{q\gamma_{k-1}^{\mathrm{raw}},\gamma_{\max}/r_\beta\}$.
SHB-PS uses this proposal as an upper bound on its raw Polyak
step size, while SHB-ALS uses it to initialize backtracking
with decay factor $\omega=0.9$.

\paragraph{Baseline methods.}
We mainly compare against SPS~\citep{SPS} and SLS~\citep{sls} without momentum, as well as fixed-step SHB.
For both SPS and SLS, we use $c=0.1$ and $\gamma_{\max}=1$.
Both baselines use the same smooth reset strategy.
For fixed-step SHB, we set $\beta=0.9$ and sweep
$\gamma\in\{10^{-4},10^{-3},10^{-2},10^{-1}\}$.

\subsection{Convex logistic regression on LIBSVM}
\label{appendix:numerical_logistic}

Following the unregularized logistic-regression protocol of
\citet{sebbouh2021almost}, we evaluate the methods on eight binary
classification datasets from LIBSVM~\citep{chang2011libsvm}:
\texttt{a9a}, \texttt{Cod-RNA}, \texttt{IJCNN1},
\texttt{Mushrooms}, \texttt{Phishing}, \texttt{real-sim},
\texttt{SVMguide1}, and \texttt{w8a}.
Each example is normalized to have unit Euclidean norm.
We use a mini-batch size of $128$ and run each method for $20{,}000$
stochastic iterations.
For fixed-step SHB, we use the same four candidate step sizes across all
datasets.

As shown in Figure~\ref{fig:svm_numerical}, SHB-PS and SHB-ALS generally reduce the training loss faster than their momentum-free counterparts.
The advantage is particularly pronounced on \texttt{Mushrooms} and \texttt{real-sim}, where the momentum variants achieve substantially lower losses within the given iteration budget.
These results demonstrate the empirical benefit of incorporating momentum into adaptive step-size methods on these logistic-regression tasks. 
\begin{figure}[t]
    \centering
    \includegraphics[width=\textwidth]{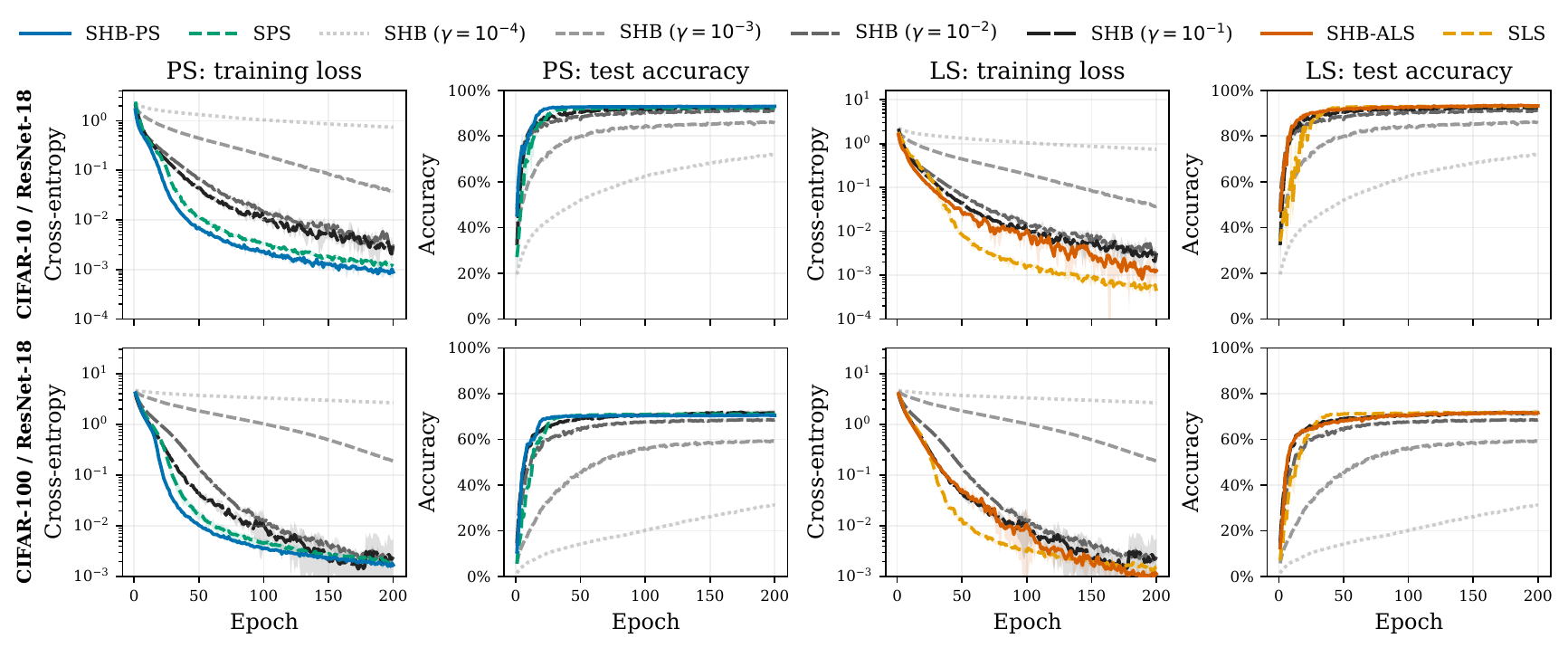}
    \caption{ResNet-18 training on CIFAR-10 (top) and
    CIFAR-100 (bottom). The first two columns compare the
    PS methods, and the last two compare the LS methods.
    We report training cross-entropy and test accuracy
    versus epochs.}
    \label{fig:resnet_numerical}
    \vspace{-1em}
\end{figure}

\subsection{Non-convex deep-network training}
\label{appendix:numerical_deep}

We evaluate SHB-PS and SHB-ALS on image-classification tasks
using ResNet-18~\citep{he2016deep} with BatchNorm on CIFAR-10
and CIFAR-100~\citep{krizhevsky2009learning}.
We train each model for 200 epochs with minibatches of size
256 and cross-entropy loss.
Training images receive random crops with four pixels of
padding and random horizontal flips.
For these experiments, we report the mean and $95\%$
confidence interval over three random seeds.

Figure~\ref{fig:resnet_numerical} shows that SHB-PS reduces
the training loss faster than SPS on both datasets and
reaches high test accuracy earlier.
For the line-search methods, SLS generally achieves lower
training loss, particularly on CIFAR-10, while SHB-ALS
attains comparable final test accuracy.
Both momentum variants also achieve lower
training losses than SHB with fixed step sizes.
Overall, these results demonstrate the practical effectiveness
of SHB-PS and SHB-ALS on deep-network training tasks, while
indicating that the benefits of momentum depend on the
step-size rules.